\documentclass[sn-mathphys]{sn-jnl}

\usepackage{makecell}
\usepackage{threeparttable}
\usepackage{multirow}

\usepackage{color}
\definecolor{turquoise}{cmyk}{0.65,0,0.1,0.3}
\definecolor{purple}{rgb}{0.65,0,0.65}
\definecolor{dark_green}{rgb}{0, 0.5, 0}
\definecolor{orange}{rgb}{0.8, 0.6, 0.2}
\definecolor{red}{rgb}{0.8, 0.2, 0.2}
\definecolor{darkred}{rgb}{0.6, 0.1, 0.05}
\definecolor{blueish}{rgb}{0.0, 0.3, .6}
\definecolor{light_gray}{rgb}{0.7, 0.7, .7}
\definecolor{pink}{rgb}{1, 0, 1}
\definecolor{greyblue}{rgb}{0.25, 0.25, 1}

\newcommand{\gan}[1]{{\color{black}{#1}}}

\newcommand{\ws}[1]{{#1}}
\newcommand{\yz}[1]{{#1}}
\newcommand{\mq}[1]{{#1}}
\newcommand{\WS}[1]{{\color{black}#1}}

\newcommand{\ournetwork}{{LineMarkNet}}

\jyear{2021}%

\theoremstyle{thmstyleone}%
\theoremstyle{thmstyletwo}%

\theoremstyle{thmstylethree}%

\begin{document}

\title[LineMarkNet]{LineMarkNet: Line Landmark Detection for Valet Parking}


\author[1,2]{\fnm{Zizhang} \sur{Wu}}

\author[2]{\fnm{Yuanzhu} \sur{Gan}}
\author[2,3]{\fnm{Tianhao}  \sur{Xu}}

\author*[2]{\fnm{Rui} \sur{Tang}}\email{rui.tang@zongmutech.com}

\author*[1]{\fnm{Jian} 
\sur{Pu}}\email{jianpu@fudan.edu.cn}

\affil[1]{\orgdiv{Institute of Science and Technology for Brain-Inspired Intelligence}, \orgname{Fudan University}, \orgaddress{\street{No. 220, Handan Road, Yangpu Area}, \city{Shanghai}, \postcode{200433}, \state{Shanghai}, \country{China}}}

\affil[2]{\orgdiv{Computer Vision Perception Department}, \orgname{ZongMu Technology}, \orgaddress{\street{Lane 55, Chuanhe Road, Pudong New Area}, \city{Shanghai}, \postcode{201203}, \state{Shanghai}, \country{China}}}

\affil[3]{\orgdiv{Faculty of Electrical Engineering, Information Technology, Physics}, \orgname{Technical University of Braunschweig}, \orgaddress{\street{Universitätsplatz 2}, \city{Braunschweig}, \postcode{38106}, \state{Lower Saxony}, \country{Germany}}}



\abstract{We aim for accurate and efficient line landmark detection for valet parking, which is a long-standing yet unsolved problem in autonomous driving. 
To this end, we present a deep line landmark detection system where we carefully design the modules to be lightweight. 
Specifically, we first empirically design four general line landmarks including three physical lines and one novel mental line. The four line landmarks are effective for valet parking.   
We then develop a deep network (LineMarkNet) to detect line landmarks from surround-view cameras where we, via the pre-calibrated homography, fuse context from four separate cameras into the unified bird-eye-view (BEV) space\gan{, specifically we fuse the surround-view features and BEV features, then employ the multi-task decoder to detect multiple line landmarks where we apply the center-based strategy for object detection task, and design our graph transformer to enhance the vision transformer with hierarchical level graph reasoning for semantic segmentation task.} 
At last, we further parameterize the detected line landmarks (e.g., intercept-slope form) whereby a novel \ws{filtering backend} incorporates temporal and multi-view consistency to achieve smooth and stable detection. 
Moreover, we annotate a large-scale dataset to validate our method. 
Experimental results show that our framework achieves \gan{the enhanced performance compared with several line detection methods and validate the multi-task network's efficiency about} the real-time line landmark detection on the Qualcomm 820A platform while meantime keeps superior accuracy, \gan{with our deep line landmark detection system.} }




\maketitle

\section{Introduction}\label{sec1}

Valet parking, aiming at driving the vehicle into the drop-off area such as parking slots, is an important and challenging task in autonomous driving \cite{song2019apollocar,zhou2020joint,wu2020motionnet}. 
With the advances of deep learning, valet parking achieves significant progress but is still bottlenecked by the drop-off area's highly complex environment resulting in inaccurate perception. 
In this case, different from autonomous driving in relatively simpler scenarios like \WS{highways and urban areas}, valet parking faces more challenges in terms of perception.

Many methods have been proposed to improve the perception of valet parking \cite{lee2017real, zhou2020joint,wu2020motionnet}. 
Among those works, the detection of traffic landmarks is one of the most effective ways \cite{Lee2022robust, hu2022sim, chen2020collaborative,wu2020psdet}. 
For instance, PSDet\cite{wu2020psdet} achieves impressive parking slot detection by recognizing a variety of landmarks around the empty slots (e.g., four corners of a rectangular parking slot.). 
In particular, line landmarks are essential for valet parking, i.e., the traffic signs that are typically long lines. In some sense, line landmark is one of the most informative traffic signs, which dominate the information source in order to park vehicles. Even as human beings, while parking cars in parking lots, we analyze and follow the line landmarks (e.g., lane line) to recognize the drivable areas and then drive the vehicles safely to the drop-off area. 
\gan{Thus, it's crucial for autonomous driving to take full advantage of these informative traffic signs, especially in parking scenes with various parking cases under different illumination conditions.}

Nevertheless, existing works are still limited by the inability to accurately detect a variety of line landmarks \cite{qin2022ultra, wang2022keypoint, feng2022rethinking,zheng2022clrnet}. 
The key challenge is that -- compared with landmarks of other shapes such as circles, points, and rectangles -- line landmarks are typically long and ambiguous (two different line landmarks could have a similar visual appearance). 
Thus, accurate detection of line landmarks requires \WS{a long-range} context which, however, is not trivial. 
On the other hand, due to ambiguous appearance, line landmarks are also ambiguous in terms of definition which results in unclear human annotations \cite{pauls2021automatic,wang2017landmarks}. 
When collecting labels for a dataset, annotators, also learners, don't have paradigms to follow and thus give error-prone labels. 
Furthermore, deep models may fail to converge without a clear definition of labels due to the ambiguous labels in training data \cite{liu2021condlanenet,tabelini2021keep}.

\begin{figure}[!t]
    \centering
    \includegraphics[width=1\linewidth]{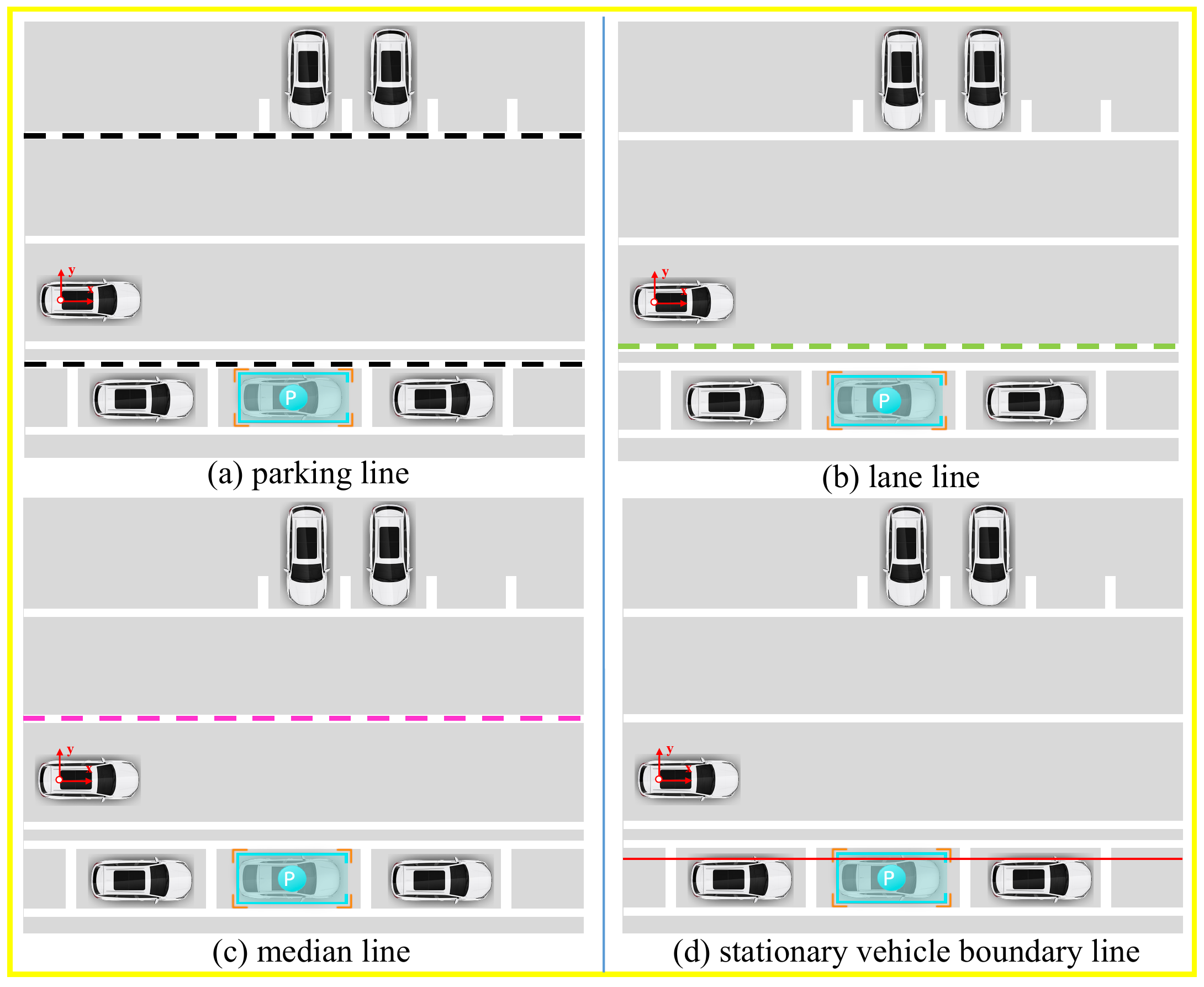}
    \caption{
    \textbf{\ws{Four} Line landmark detection for valet parking}:
    \ws{Inspired by empirical observations from industrial applications}, we summarize four general line landmarks which are essential for valet parking. 
    Specifically, we propose three physical line landmarks (a: parking line, b: lane line, c: median line) and one mental line (d: stationary car boundary line). 
    \ws{Note our mental line is a virtual line connecting stationary cars (parking in the slots) and serves as the tight boundary between drivable and non-drivable areas. The other three physical line landmarks play a key role in navigation, particularly in the parking environments.    
    We further elaborate on the motivation of the proposed four line landmark in Sec. \ref{sec:def_line}.          
    } 
    }
    \label{scene}
\end{figure}

\WS{In light of this}, we empirically hand-craft four common and essential types of line landmarks by observing a variety of possible environments for valet parking.  
Specifically, 
we collect and observe a large-scale dataset for valet parking, assuring that the summarized four line landmarks are representative.
As shown in Fig. \ref{scene}, the four line landmarks consist of three physical line landmarks (visible lines such as lane lines) and one mental line landmarks (i.e., a virtual line that connects the stationary vehicles parking along lanes). It's noteworthy that our novel mental line landmarks aim to define the tight boundary between drivable and non-drivable areas.      
Moreover, we annotate the collected dataset, allowing us to train our network for line landmark detection.  

Nevertheless, it's still challenging to incorporate long-range context for the detection of long-line landmarks. One key challenge from the hardware aspect is that, due to the limited field of view, the observations of long line landmarks via cameras are generally incomplete. 
To address this issue, we leverage the surrounded-view camera system to achieve the complete observation of line landmarks. It consists of four fisheye cameras whereby a large field of view enables line landmarks to be fully visible. 
However, as the other challenge from the software side, it's still non-trivial for deep networks to extract long-context since line landmarks distribute over multiple views. To solve this problem, we develop the LineMarkNet that aims to detect four common line landmarks from surround-view cameras. The core idea is that,  via \WS{the pre-calibrated homography} of four cameras, we transform image planes into the BEV space whereby context information from multiple views is fused. 
\gan{It adopts the multi-task architecture, where we firstly fuse the surround-view features and BEV features, then utilize the multi-task decoder (semantic segmentation and object detection) and calculates the line parameters through the followed line fitting module as shown in Fig.~\ref{fig_linemarknet_framework}.
Within the decoders, we apply the center-based strategy for object detection and design our graph transformer network to improve the semantic segmentation performance, which enhances the global and local relation modeling with hierarchical level graph reasoning.
}

Our experiments show that LineMarkNet achieves impressive performance. However, we observe that the detected line landmarks are still unsurprisingly noisy due to occlusion. Note that while the noisy output is not surprising, it's not satisfying for valet parking where high precision is required.
Thus, to mitigate the noise, we refine the initial output of LineMarkNet through a novel \ws{filtering} backend. The idea of the \ws{filtering} backend is to enforce multi-view and temporal consistency where we encourage the line landmarks to be consistent across multiple fisheye cameras and adjacent frames. Our experiments show that our backend effectively removes noises and achieves smooth and stable line landmark detection.             

We observe in our experiments that our system achieves accurate line landmark detection. Moreover, we carefully design modules of our system to be lightweight -- both memory and computationally efficient. This enables our line detection system to run in real-time on the Qualcomm 820A platform.

We simply summarize our contributions as:
\begin{itemize}
    \item we define four general and informative line landmarks including three physical lines and one novel mental line. The four line landmarks are effective and essential for valet parking.
    \item we propose LineMarkNet to detect line landmarks from the surround-view camera system,
    \gan{which adopts the multi-task architecture. Within it, we fuse the surround-view features and BEV features to the improved multi-task decoder (semantic segmentation and object detection) and calculate the line parameters through the following line fitting module.}
    \item we annotate a large-scale dataset which allows us to train our LineMarkNet, \gan{which contains plenty of the long and ambiguous line landmarks in the valet parking scenes.}
    \item we further propose a novel \ws{filtering} backend to refine the initial output of LineMarkNet where we filter out incorrect line landmarks by enforcing multi-view and temporal consistency.
    \item our line mark detection system achieves superior accuracy while being real-time, \gan{compared with the previous line detection approaches.}  
\end{itemize}
\gan{The other parts of the paper are organized as follows. The second section of the paper introduces the related methods of the paper. The third section introduces the details of the method in this paper. The fourth section describes the experimental comparison and analysis of the method proposed in this paper. Finally, the fifth section is about the overall summary of the paper.}

\section{Related Works}\label{sec2}
We discuss related works that utilize surround-view cameras, review the method dealing with surround-view images, and briefly summarize the methods for line landmark detection as below. 

\subsection{Surround-view camera system}
Thanks to \WS{its large field-of-view} \cite{2018IVreal, wu2021disentangling}, \WS{the surround-view camera system is a very popular imaging system} for visual perception \cite{2018IROSend,2021arxivomnidet,zhang2021avp}.      
In particular, the frameworks for autonomous driving \cite{kumar2020fisheyedistancenet, yahiaoui2019fisheyemodnet,2020IROSAccuratelow, qin2020avp} are typical of the surround-view camera system that consists of four fisheye cameras mounted on the ego-vehicle \cite{ouyang2020online,wang2020reliable}. 
It's thus capable of capturing all objects nearby the ego-vehicle, allowing for reliable visual perception.
In this work, we utilize a similar camera system in our line landmark detection system.       
Note that our focus is on the perception of valet parking, which is more challenging than methods for autonomous driving.   

\subsection{Deep frameworks for surround-view camera system}
Recently, lots of deep methods have been proposed for surround-view camera systems with a variety of target tasks such as semantic segmentation \cite{2018ECCVWsemantic,kumar2021syndistnet,ye2020universal}, distance estimation \cite{kumar2020fisheyedistancenet,zaffaroni2019estimation,komatsu2020360} and object detection \cite{yahiaoui2019fisheyemodnet,plaut2020monocular,li2020fisheyedet}. 
The most relevant works are the methods for semantic segmentation and detection as our system utilizes these two tasks for line detection.       
Note that one of the most desiderata for these methods is the strategy to fuse information from multiple viewpoints.    
To address this, existing works \cite{wu2020psdet} typically fuse deep features in the BEV space. 
In general, deep features are learned from multiple views individually via the siamese network \cite{wu2020psdet}.    
We follow a similar strategy to fuse information from multiple views, with modifications (e.g., line-fitting module) specifically for line landmark detection.  
Moreover, in our system, we have the \ws{filtering} backend that further compresses the potential noise or outliers during the fusion process.   

\subsection{Line landmarks detection}
Line landmarks are kind of semantic lines, existing in many tasks \cite{2020xiao,hu2019mapping}.
Note that we exclude the discussion about the low-level line feature \cite{qin2018vins} as it's quite different from our line landmarks by definition.
In some sense, by defining four line landmarks, we endow line features with the semantics to be learned using deep networks. 
In this context, the most relevant task is the lane detection \gan{\cite{2020Marzonggui, 2020xiao, qin2020ultra,huang2019dmpr,Lee2022robust,hu2022sim,wang2022keypoint,feng2022rethinking}} where the lane line is modeled as the line landmarks. 
So far, previous works \gan{\cite{2020xiao,2018Song,qin2020ultra, zheng2022clrnet, liu2021condlanenet, tabelini2021keep}} have achieved promising lane detection. 
However, more types of line landmarks exist in the environment for valet parking whereas, to our best knowledge, previous works rarely investigate the general line landmark detection as in our benchmark.
In this case, our method bridges the gap between effective deep methods and missing pipelines for line landmark detection in valet parking.

\section{Methods}

\begin{figure*}[!t]
    \centering
    \includegraphics[width=1.0\textwidth]{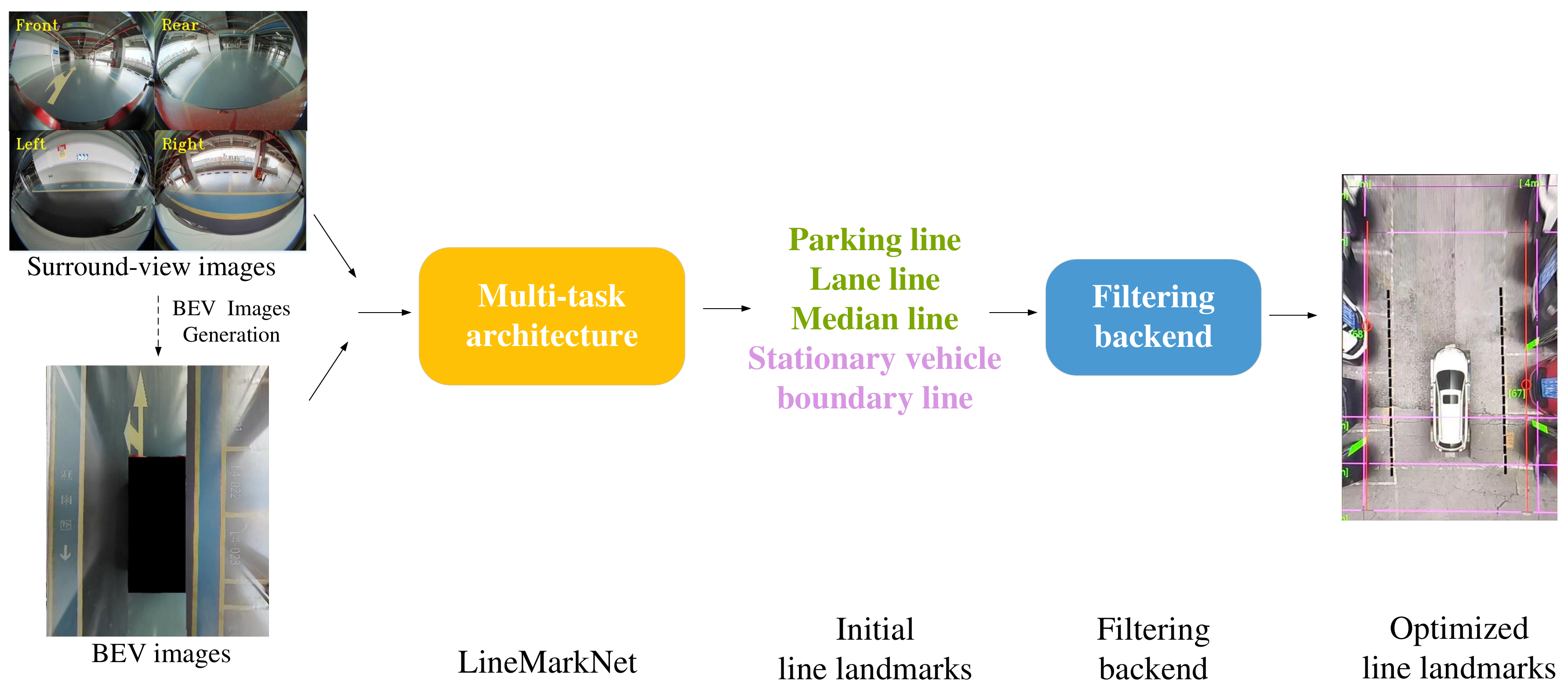}
    \caption{\textbf{Line landmark detection system -- }
    Our system \ws{consists} of the \textit{LineMarkNet} -- a multi-task architecture that detects multiple line landmarks from the surround-view cameras -- and a novel \ws{\textit{filtering backend}} to removes detection errors in the output of lineMarksNet. Our system effectively \ws{detects} the line landmarks detection and meanwhile is real-time.  
    }
    \label{mult-task}
\end{figure*}

We introduce a deep line landmark detection system for valet parking. 
As illustrated in Fig. \ref{mult-task}, the system consists of a LineMarkNet that detects line landmarks from surround-view cameras,  and an efficient filtering backend where we further enhance the results by filtering out noise and outliers in the output of LineMarkNet.      
Note that, while having two complex components,  our system is very efficient such that we achieve the real-time line landmark detection on the Qualcomm 820A platform.

Specifically, we detect line landmarks from surround-view cameras. 
As shown in Fig. \ref{ipm}, the surround-view camera system consists of four fisheye cameras with a large field of view whereby the long line landmarks are fully visible in four images. 
To detect line landmarks from \WS{surrounding views}, we propose \ournetwork{} as illustrated in Fig. \ref{fig_linemarknet_framework} where we employ a multi-task architecture to detect multiple line landmarks.   
Long-range contextual information is the key to detecting long-line landmarks.
Thus, we transform the information extracted in 2D image planes into the BEV space via calibrated homography \cite{zhang2000flexible}. 
In some sense, we detect line landmarks in BEV space.  

Additionally, we propose the \ws{filtering} backend to further refine the results from \ournetwork{}. Although the initial detection from \ournetwork{} surpasses naive baselines, the line landmark detection is still noisy whereas valet parking requires very high precision to order keep the safety of driving. 
To mitigate the issue, we propose to optimize detection by enforcing the multi-view and temporal consistency. 
Specifically, our \ws{filtering} backend encourages the line landmarks detection to be consistent across different views and adjacent frames. As a result, our system achieves stable and consistent line landmark detection which enables practical applications in valet parking.          

We will detail our method below.

\subsection{Definition of line landmarks}
\label{sec:def_line}
\ws{
Our empirical experiences gained in industrial software development mainly motivate the four line landmarks whereas our four line landmarks suffice to navigate vehicles in part lots.  
Interestingly, although a variety of line landmarks exist in the complex parking lots, merely a few common line landmarks 
is sufficient for valet parking.              
The reason is that line landmarks are human-designed for certain goals. In particular, one of the most important goals is to navigate vehicles. 
For instance, the lane marks,  by design, aim to separate different lanes whereby vehicles are \ws{navigated} to drive in the legal lanes. 
In essence, all our line landmarks aim to navigate vehicles in the parking environments except for the minor differences in functionality -- e.g, navigating in different areas, leading to our four line landmarks.   
Note that, despite the possibility of neglecting useful line landmarks, it's trivial to adapt our line landmark detection system to new line landmarks, allowing our system to generalize into new types of parking lots. 
}

As shown in Fig. \ref{scene}, we summarize four types of line landmarks. And we further briefly describe the functions and pattern as       
\begin{itemize}
    \item \textbf{Parking line} (Fig. \ref{scene} a) is the landmark that separates drivable lanes and parking slots, \ws{essential for vehicle to localize the target parking slot}. The parking lines are typically long and continuous (i.e., the landmark is a straight line.).     
    \item \textbf{Lane line} (Fig. \ref{scene} b) separates multiple drivable lanes, \ws{enable vehicle to drive on a certain lane}. In general, it's the line landmarks closest to the vehicle. And it can be continuous or discrete (i.e., the landmark is composed of a set of line segments.).     
    \item \textbf{Median line} (Fig. \ref{scene} c) differentiate between lanes of two directions. \ws{It's critical to avoid collisions between vehicles from different directions. As illustrated in Fig. \ref{scene}, }it's generally a long and straight line.  
    \item \textbf{Stationary vehicle boundary line} (Fig. \ref{scene} d) is a mental line landmark -- it doesn't physically exists. We define it as the line that connects stationary vehicles (i.e., vehicles parked in slots). We connect the {mid points} of stationary vehicles closest to the lane. 
    Thus, our novel mental line serves as the tight boundary between drivable and non-drivable areas. The motivation of this mental line is very straightforward since as human-being, we mentally draw some auxiliary lines to help driving.          
\end{itemize}
The line landmarks of the same type share similarity in terms of visual pattern. Thus, we easily collect labels according to the visual pattern which enables the training of our pipeline to detect line landmarks. 
\subsection{The Framework of LineMarkNet}

\begin{figure*}[!t]
    \centering
    \includegraphics[width=1.0\textwidth]{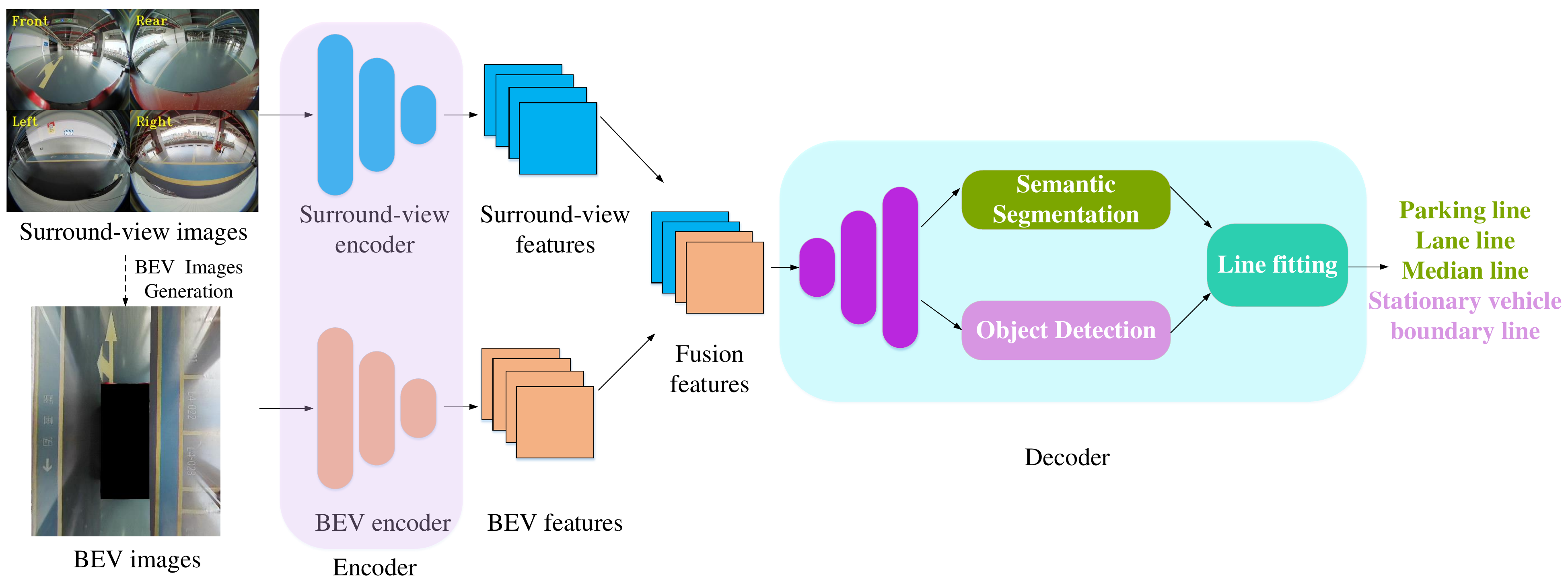}
    \caption{\textbf{The framework of LineMarkNet} --  The network takes as input the surround-view fisheye images, which allows for the line landmark detection with the large field of view. Specifically, we fuse contextual information in BEV space and performance multiple tasks where our system perceives line landmarks via tasks of semantic segmentation and object detection and calculates the line parameters through the followed line fitting module.  
    }
    \label{fig_linemarknet_framework}
\end{figure*}

We introduce the LineMarkNet where we detect line landmarks from surround-view images.
To fuse information in different views (BEV and surrounding views), we transform the information extracted in 2D image planes into the BEV space. 
The multi-task decoders make full use of the fusion features to detect and vectorize the four general and informative line landmarks. Finally, the efficient \yz{filtering backend} enhances the line landmarks to be more smooth and \WS{more stable}.
Fig. \ref{mult-task} shows the overall structure of our proposed system.

\subsubsection{BEV Images Generation}
Our system needs to obtain BEV images of the surrounding environment. 
Generally, the camera’s optical axis has a certain inclination angle with the ground, so we need to conduct the perspective transformation to obtain the BEV images. 
We choose the center of the front camera as the origin of the world coordinate system.
We change the perspective of the images according to the linear transform of the homography matrix \cite{zhang2000flexible}. We can compute the coordinate ($x$, $y$) in the image coordinate system from the coordinate ($X$, $Y$, $Z$) in the world coordinate system using the following homography transformation:

\begin{align}
\label{equa4_9}
&x = \frac{G_1X + G_2Y + G_3Z + G_4}{G_9X +G_{10}Y + G_{11}Z +1} 
\\
\label{equa4_10}
&y = \frac{G_5X + G_6Y + G_7Z + G_8}{G_9X +G_{10}Y + G_{11}Z +1} 
\end{align}

where $G_1$, $G_2$, $G_3$, ..., $G_{11}$ are unknown parameters. When projecting the object to the horizontal plane, $Z$ in the world coordinate system becomes zero.
So we select four sets of corresponding coordinate points in the world coordinate system and the image coordinate system: ($X_0$,$Y_0$), ($x_0$,$y_0$); ($X_1$,$Y_1$), ($x_1$,$y_1$); ($X_2$,$Y_2$), ($x_2$,$y_2$); ($X_3$,$Y_3$), ($x_3$,$y_3$), and substitute them into Equation (\ref{equa4_9}) and (\ref{equa4_10}) to solve the parameters (from $G_1$ to $G_8$), and complete the transformation process.

\begin{figure}[!b]
    \centering
    \includegraphics[width=1.0\textwidth]{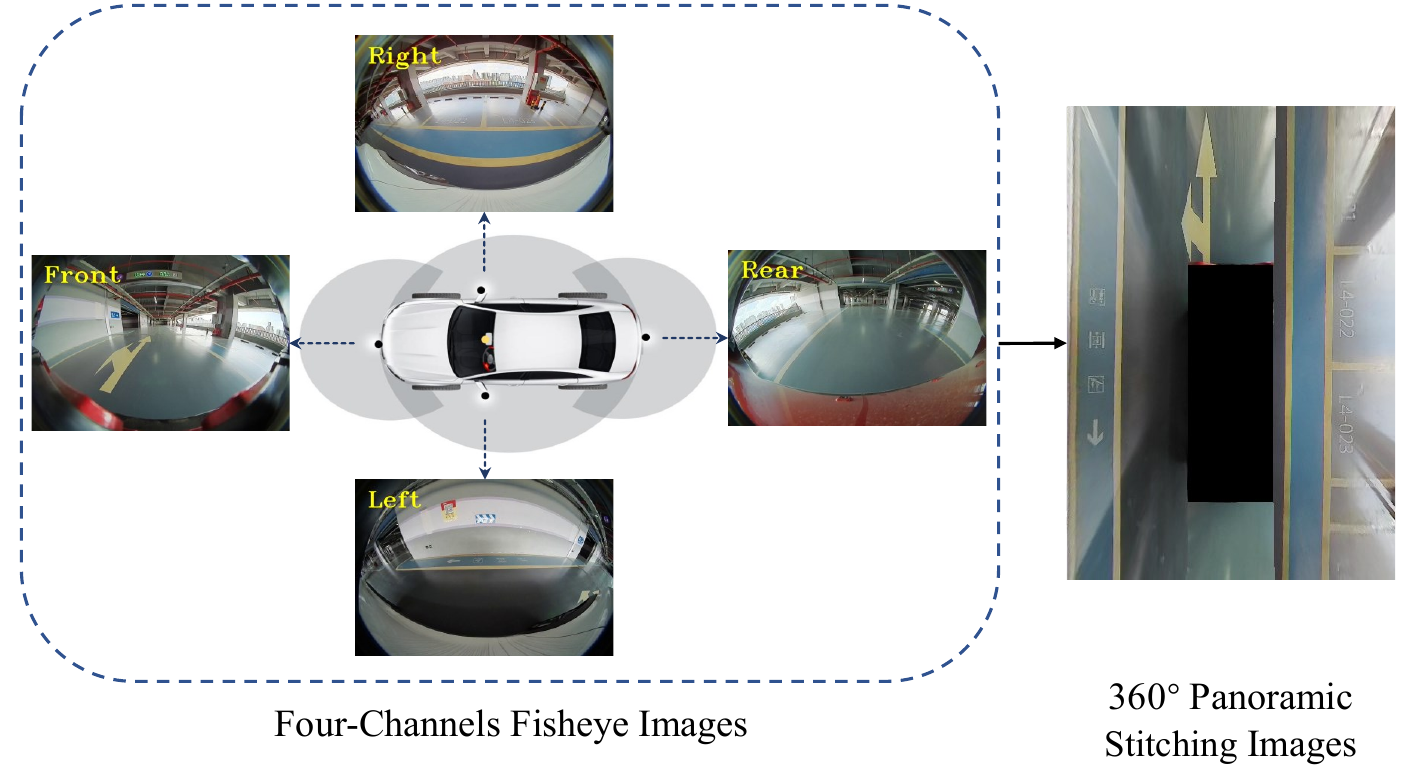}
    \caption{Illustration of four surround-view cameras and the synthetic BEV image from four surround-view camera images. Best view in color and zoom in.}
    \label{ipm}
\end{figure}

Fig. \ref{ipm} presents the illustration of four surround-view cameras and the synthetic BEV image from four surround-view camera images.
After obtaining the parameters from $G_1$ to $G_8$, we can capture the physical coordinates from pixel coordinates through the homography transformation.

\begin{figure*}[!t]
    \centering
    \includegraphics[width=1.0\textwidth]{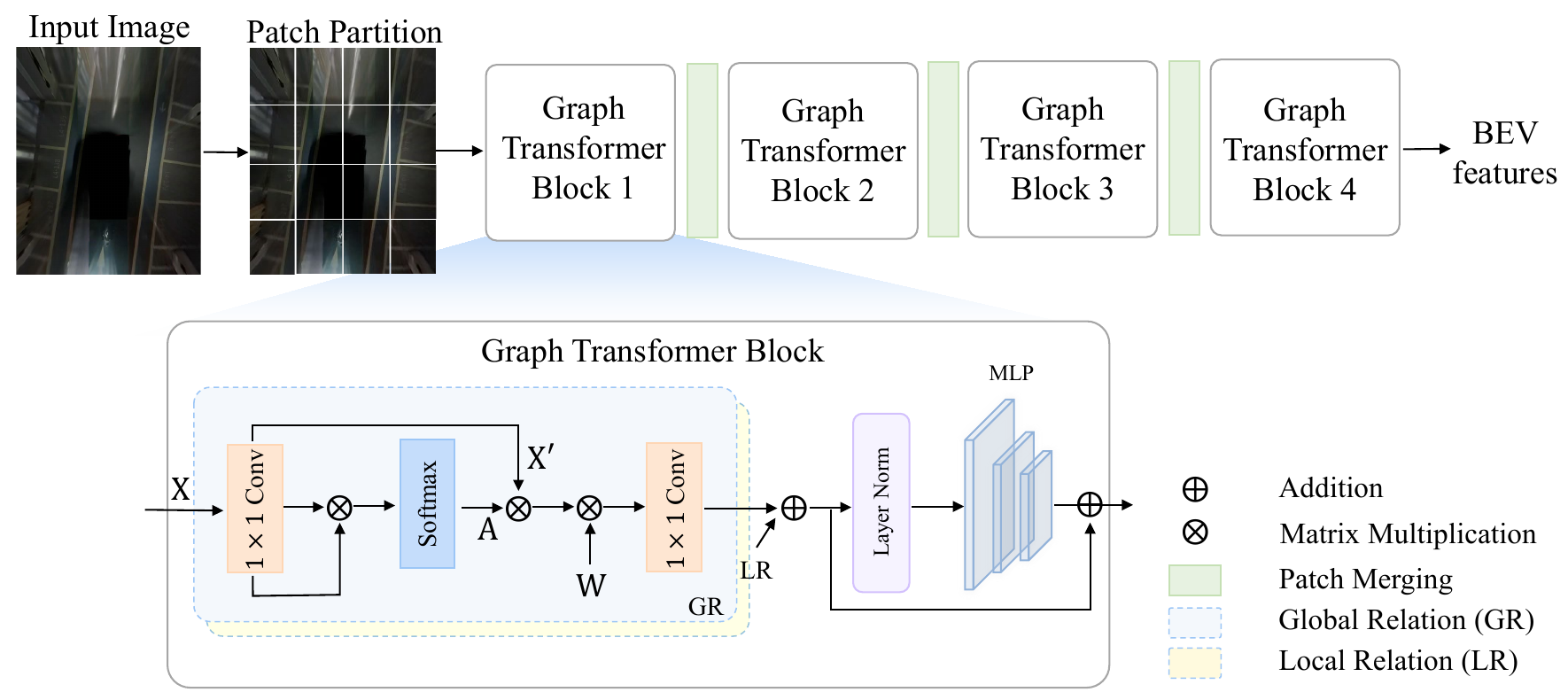}
    \caption{\yz{The overall structure of our BEV encoder. 
    The BEV encoder exploits graph transformer, which enhances the vision transformer with hierarchical level graph reasoning, including global and local relation modeling}. Best view in color and zoom in.}
    \label{net_frame}
\end{figure*}

Our system is a multi-task architecture where various tasks share the image features, and different decoders are designed for specific tasks, as shown in Fig. \ref{fig_linemarknet_framework}.

\subsubsection{Encoder}
The encoder consists of the surround-view encoder and BEV encoder, extracting informative image features from surround-view images and BEV images.      

\paragraph{Surround-view encoder (CNNs)}
\yz{Our surround-view encoder takes as input the four surround-view images and outputs the surround-view features.
\mq{We concatenate surround-view images as batched input -- a tensor with the batch size of 4. In other words, our encoder acts as a siamese network that is weight-sharing among four views. The tensor shape indicates $[4,3,H_{S},W_{S}]$, where $H_{S}$ and $W_{S}$ mean the surround-view images' height and width.
We adopt the standard DLA-34 \cite{dla} as the surround-view encoder, for a better speed-accuracy trade-off.
Finally, the surround-view encoder produces surround-view features: $[4,N_{S},h_{m},w_{m}]$, where $N_{S}$ denotes the channel of feature maps, $h_{m},w_{m}$ mean feature maps' height and width.
}
}

\paragraph{BEV image encoder (graph transformer)}
\yz{
Unlike surround-view images, \mq{BEV images consist of long-range contexts and need the feature extractor with a larger receptive field}. 
In this case, we propose to learn the BEV features with the graph transformer (i.e., our BEV image encoder), which is an enhanced vision transformer with better \mq{long-range contextual information}. 

\mq{
We firstly state the input and output.
Four surround-view images can create one BEV image, denoting $[1,3,H_{B},W_{B}]$. 
To match the 4 batch-size surround-view features when feature fusing, we need to copy the BEV image 4 times to form $[4,3,H_{B},W_{B}]$ as the BEV encoder's inputs. 
Then the BEV encoder can output $[4,N_{B},h_{m},w_{m}]$ features, with the same shape with surround-view features.

Below we introduce our BEV encoder, as shown in Figure \ref{net_frame}.
}
We divide one image into patches and define one window composed of some (like 7$\times$7) patches. 
In other words, one image can be divided into some windows and one window includes 7$\times$7 patches. 
Firstly, we propose the Global Relation (GR) module to reflect the global modeling among different windows. Specifically, we consider each window as one graph node and use the efficient graph relation network to build the relation among the graph nodes. 
\mq{Then we update these nodes using graph convolution \cite{bruna2014spectral}.
The GR module can obtain the coarse relation of the objects in each image.}
Afterward, we conduct the Local Relation (LR) module to build relations inside each window,\mq{ meaning the patch-level relations.} 
Considering each patch as a node, the module builds the graph relation network to model the local relation within each window. 

\mq{
The original Swin \cite{liu2021swin} employs the shifted window technique to strengthen windows' connection, but this interaction with other windows is unfree with fixed orientation and fixed offset, where the long-distance relationships between windows/patches are not thoroughly explored.
We adopt graph convolutions to establish relationships between windows and patches inside each window, which enhances the block to address the above issues.
}Actually, both the GR module and LR module can be regarded as one kind of self-attention mechanism.

It's noteworthy that the graph transformer block can be implemented by two $1\times1$ Conv, a normalization function (Softmax) and other basic elements (\mq{LayerNorm} and Multilayer Perceptron in Swin \cite{liu2021swin}). 
Specifically, we first use $1\times 1$ Conv to reduce the computational complexity within the block (saying X becomes X$^{\prime}$). Similarly, another $1\times 1$ Conv is realized to resume the channel dimension. 
To obtain the relation matrix A, we use a matrix multiplication and a softmax function to normalize the results. 
W is the learnable parameter of graph convolution. 
So the graph node update process can be accomplished by two matrix multiplication simply: A$\otimes$X$^{\prime}$$\otimes$W. $\otimes$ means the matrix multiplication.

The LR module's node update is the same as the GR's, but with different learnable parameters. At last, we fuse (add) these two relation modeling modules to the following \mq{LayerNorm} and MLP modules. In the end, the graph transformer can extract more robust BEV features.
}
After concatenation between the surround-view features and BEV features, the produced fusion features can contain long-range contextual information.

\subsubsection{Decoder}



Multi-task decoders contain the semantic segmentation task and object detection task, which extract line landmarks' pixels and perform vectorized outputs after different post-processing.
\begin{figure}[!t]
    \centering
    \includegraphics[width=1.0\textwidth]{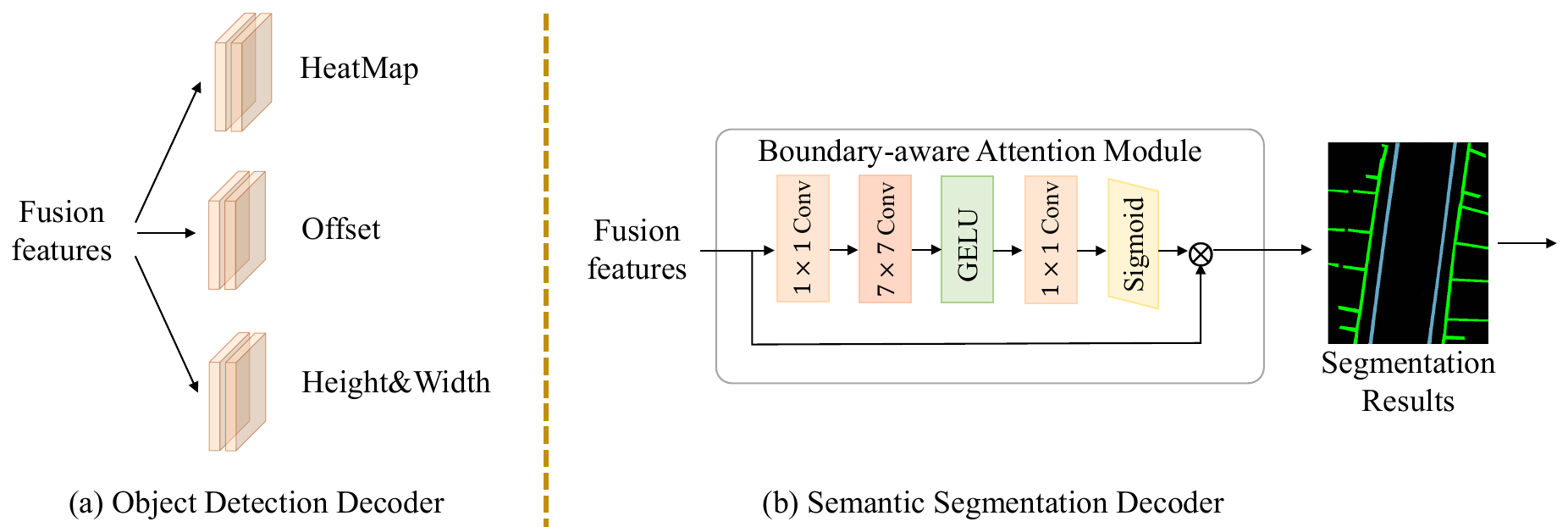}
    \caption{\mq{The overall structure of our decoder, including object detection decoder and semantic segmentation decoder.
    Best view in color and zoom in.}
    }
    \label{ss_decoder}
\end{figure}

\paragraph{Semantic segmentation decoder}

\mq{
To better optimize the boundaries of objects, we add a boundary-aware attention module in the decoder.
We perceive the relation between local pixels, based on which we learn to obtain the boundary-aware attention coefficients of each surrounding pixel, as shown in Fig. \ref{ss_decoder}(b).
Concretely, we first use two $1\times 1$ Conv to reduce and resume the channel dimension.
Then we adopt the local modeling function to learn the local relation, where the pixels around the boundary of the object are adjusted.
Here we simply use $7\times 7$ Conv to implement the local relation modeling function. 
Besides, we add the non-linear function GELU inside the boundary-aware module to learn more robust relations.
Finally, we choose Sigmoid as our normalization layer.
The boundary-aware attention module emphasizes on boundary-aware attention with \WS{local relations}, which results in a better segmentation performance.
}



\begin{figure}[!b]
    \centering
    \includegraphics[width=1.0\textwidth]{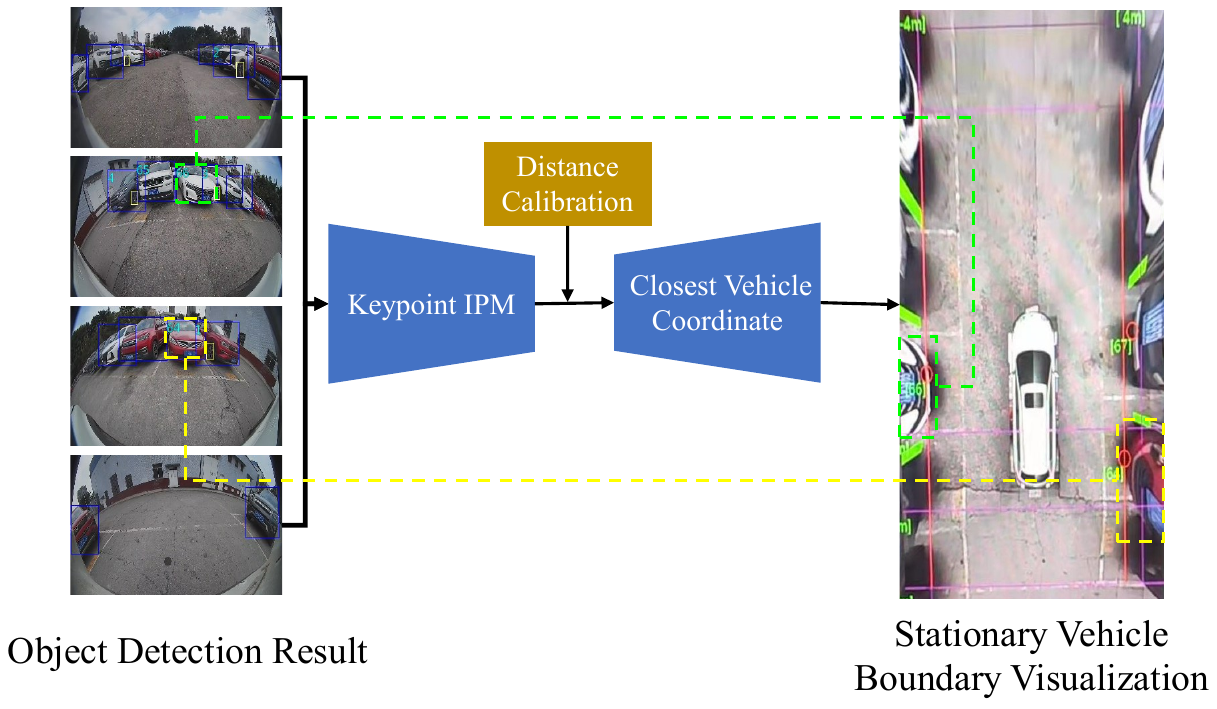}
    \caption{The structure of stationary vehicle boundary generation method. Best view in color and zoom in.}
    \label{Stationary_Vehicle}
\end{figure}

\paragraph{Object detection decoder}
The stationary vehicle boundary line is important for helping automatic parking vehicles to determine the drivable area more accurately. 
\yz{
To generate it, we need to detect and locate surrounding vehicles from the fisheye images, then compute their key points, as shown in Fig. \ref{Stationary_Vehicle}.

So we perform 2D object detection on fisheye images and adopt the same decoder as CenterNet \cite{2019arxivobjects} for a trade-off between real-time performance and accuracy,\mq{ as shown in Fig. \ref{ss_decoder}(a).
The object detection decoder regresses objects center points' heatmap and offsets, as well as object's height and width, to fulfill the objects' correct location.}
3D object detection is abandoned because of the difficulty of acquiring the large scale of 3D labeled truth data and unsatisfactory real-time performance.}



\paragraph{Line fitting}

From segmentation results, we detect three kinds of line landmarks, including parking lines, lane lines, and median lines. 
\yz{The parking lines contain horizontal and longitudinal lines, where longitudinal parking lines are along the heading direction, beneficial to automatic parking controlling; while horizontal lines are perpendicular to the car's heading direction, which contains little useful information for driving and parking.}
Line detection should strive for both quality and speed.
So we propose a simple and ingenious line detection algorithm to improve performance on embedded devices.

Firstly we perform the lane line and median line's detection according to semantic segmentation results.
Then horizontal scan lines with equal intervals traverse these landmarks. 
Every horizontal line creates some intersection points with the lane line or median line. 
We count every horizontal line's intersection points' number and get the most occurrences of the intersection points' number.
We choose the horizontal lines with the most occurrences of \WS{intersection numbers} and select their intersection points to fit the lane line and median line by the least square method. 
At last, we calculate the vectorized information (slope and intercept) of the fitted line landmarks, and the rotation angle based on the longitudinal axis of the ego-vehicle. \yz{Algorithm \ref{algorithm1} shows the detailed steps.

}

\begin{algorithm}[t]
\caption{\yz{Lane line and median line’s fitting}}%
\textbf{Input:} Semantic segmentation results \\
\textbf{Output:} Vectorized information (slope and intercept) \\
\textbf{Step 1}: Applying horizontal scan lines $\{l_1,l_2,\ldots,l_n\}$ with equal intervals to traverse the semantic segmentation results.  Every horizontal line returns $\{c_1,c_2,\ldots,c_n\}$ intersection points.
\\
\textbf{Step 2}: Counting most occurrences of the intersection points' number $\{c_1,c_2,\ldots,c_n\}$, and getting $c_*$
\\
\textbf{Step 3}: Reserving the horizontal scan lines with $c_*$ intersection points': $\{{l_1}^{*},{l_2}^{*},\ldots,{l_m}^{*}\}$ and abandoning the others.
\\
\textbf{Step 4}: Obtaining the intersection points from $\{{l_1}^{*},{l_2}^{*},\ldots,{l_m}^{*}\}$. Every ${l_i}^{*}$ ($i\in \{1,2,\ldots,m\}$) can obtain $c_{*}$ intersection points: $\{{p_{i1}},{p_{i2}},\ldots,{p_{i(c_{*}-1)}},{p_{{ic_{*}}}}\}$ in sequence. 

\textbf{Step 5}: Fitting lines. Selecting $\{{p_{i1}},{p_{i2}}\}$ to fit the first line's vectorized information by the least square method, since one line could create two adjacent intersection points. Completing the $c_{*}/2$ lines' fitting using the same procedure.
\label{algorithm1}
\end{algorithm}

Next, we carry out the detection of the longitudinal parking lines.
We rotate the image by the rotation angle and traverse the parking lines longitudinally to detect longitudinal parking lines' points without the vertical lines’ disturbance. 
Then we rotate back the images and use the points after rotation to fit the longitudinal parking lines with the least square method and get the longitudinal parking lines' vectorized information.

For stationary vehicle boundary lines, after obtaining the bounding boxes of the vehicles from the object detection decoder, we regard the midpoints of the lower bottom of the vehicle boxes as the key points and acquire their pixel coordinates in the image through post-processing. 
These points are special since their coordinates in the ego-vehicle coordinate system are $P(x, y, 0)$. After obtaining the pixel coordinates of the key points, we can get these physical coordinates in the real-world coordinates through the Fisheye IPM algorithm \cite{mallot1991inverse}.

With the help of the distance calibration table, we can obtain the coordinate points of stationary vehicles closest to the ego-vehicle in the lateral direction. 
Finally, the line passing through these points and parallel to the ego-vehicles heading direction is the stationary vehicle's boundary line.
The red line on the right side of Fig. \ref{Stationary_Vehicle} shows the visualization results.

\subsection{\ws{Filtering backend}}
\ws{
Our filtering backend endows the system with the ability of refining initial line landmark detection. 
While our LineMarkNet effectively detects line landmarks, there is still a non-negligible amount of noise and outliers present in initial outputs.
The basic idea behind our novel backend is that, with a carefully designed filter, our filter backend smooths the initial results by encouraging the consistency of line landmarks in two aspects: (1) \textit{multiview consistency:} Detection from multi-view images should be consistent. and (2) \textit{temporal consistency: } Detection from temporal frames should be consistent.  
To be more specific, for the former, our backend requires consistent line landmarks detection from four fisheye cameras. 
For the latter, it encourages the temporal consistency via Kalman filtering where we estimate the inconsistency between Kalman's prediction based on previously detected line landmark and current detection. 

\mq{In some sense, our filtering is similar to existing works \cite{kalman1,kalman2,kalman3,qin2020avp}. 
However, these existing works are targeted for their specific tasks like odometry, which can’t be transferred directly to our line landmark detection.
We customize the filtering, aiming for the new problem in valet parking -- i.e., line landmarks detection.}

Our experimental results show that, by enforcing these two consistencies, our \ws{filtering backend} achieves a more stable and smooth line landmarks detection. Moreover, the \ws{filtering backend}, thanks to its lightweight nature, is very efficient and thus enables our entire line landmark detection system to be real-time.      

We will elaborate multi-view and temporal consistency as below.

\subsubsection{Multi-view consistency}
Our backend enforces the multi-view consistency by associating objects observed in multiple view cameras. 
Thus, apart from the stationary vehicle boundary line, the other three line landmarks estimated in the BEV space don't require this step. 
Nevertheless, it's noteworthy that BEV based line landmarks theoretically guarantee the multi-view consistency since multi-view information is fused before the detection. 
Therefore, to enforce the multi-view consistency for the stationary vehicle boundary line,  we associate vehicles detected from different cameras via the defined association scores.   
Specifically, we estimate the consistency of two detections from different cameras as the distance of midpoints projected to the ego-vehicle coordinate system 
whereby two stationary vehicles are associated if its' midpoints distance is smaller than 25cm.        

\subsubsection{Temporal consistency}
We base temporal consistency on the Kalman filter algorithm where, with motion equations, our backend predicts line landmarks at the current time frame.
Following the spirit of the Kalman filter, we construct the vehicle motion equation that allows us to predict the current multi-dimensional state of line landmarks. 
Specifically, the multi-dimensional state ($s$) consist of line landmarks' center point (i.e., center pixels of $c$) and parameters (i.e., intercept $\theta$ and slope $\beta$). 
Denote the predicted multi-dimensional state at time frame $t$ as 
\begin{align}
   \Tilde{s}_t = [\Tilde{c}_{t}; \Tilde{\theta}_{t}; \Tilde{\beta}_{t}] 
\end{align}
Thus, our backend, in some sense, tracks the line landmarks via multi-dimensional information. 
Note that additional dimensions would help. But we empirically observe that center points and line parameters are sufficient to achieve stable and smooth line landmarks detection. 
Moreover, the state $s$ has a small footprint which is the key reason why our system is capable of being real-time. 

In the meantime, our network detects the line landmarks at time $t$ where we obtain the state $s_t$ from \WS{the current} given surround-view images.
Accordingly, we quantify the inconsistency ($\sigma$) as   
\gan{
\begin{align}
\sigma_t = \lambda_1 \mid\mid c_{t} - \Tilde{c}_{t}\mid\mid 
+ \lambda_2 \mid\mid \theta_{t} - \Tilde{\theta}_{t}\mid\mid 
+ \lambda_3 \mid\mid \beta_{t} - \Tilde{\beta}_{t}\mid\mid 
\end{align}}
where weights $\lambda_1$, $\lambda_2$ and $\lambda_3$ adjust the contribution of different dimensions on the final inconsistency .
By doing so, our \ws{filtering backend} removes the line landmarks detected at the current frame via thresholding on the inconsistency. 
Specifically, for any line landmark, we calculate its inconsistency and remove it from detection if its inconsistency $\sigma$ is larger than a predefined threshold. In this way, our system can filter out the incorrect line landmarks that are quite different from previous detections.        

Note that, slightly different from the typical object tracking that requires the association process -- i.e., new detection needs to be associated with old detection before calculating the inconsistency, our backend takes the advantage of the unique nature of line landmarks. That being said that we mildly assume that our line landmarks, by design, have a single instance present in the given scene and thus can eliminate the association process.          
Even though, it's noteworthy that it's not nontrivial to add the association process into our backend by leveraging our multi-dimensional state.  
}

\section{Experiments}

In this section, we first detail our experimental setup. 
Then, we provide quantitative and qualitative results to validate the effectiveness of our method.  
We further demonstrate the ablation study on different components of our method. 
The results show that our method achieves effective yet efficient line landmark detection.

\subsection{Experimental Setup}

\subsubsection{Dataset} 
To validate our method, we collect a large-scale dataset for line landmark detection in valet parking. 
The dataset consists of 140k training samples and 400k testing samples.
It's noteworthy that, to handle a variety of parking scenarios, we collect our data from over 400 parking lots distributed in different places.
\begin{figure}[th!]
    \centering
    \includegraphics[width=0.95\linewidth]{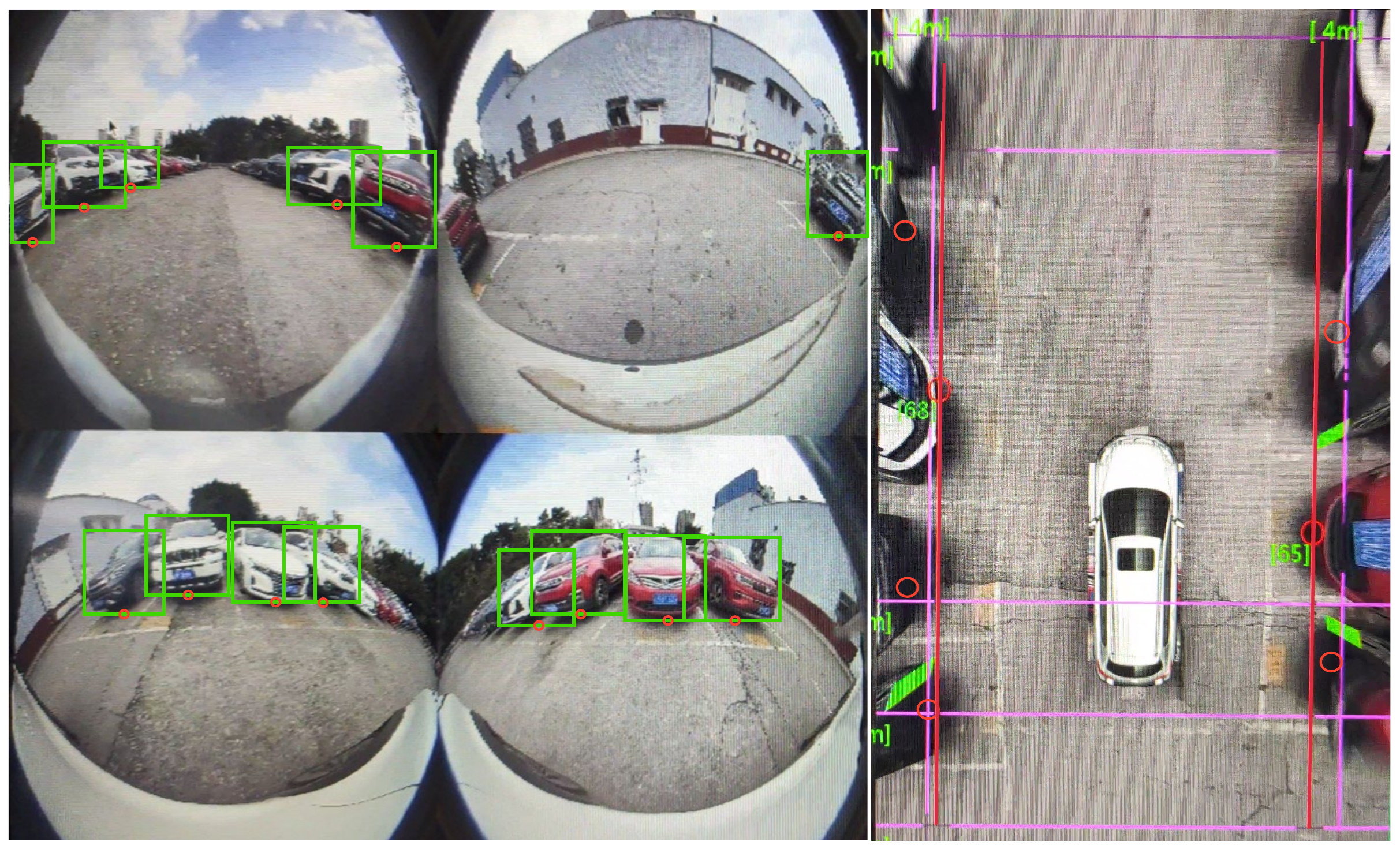}
    \caption{\textbf{An example of stationary vehicles in surround-view images' labelling}
    We annotate the bounding box for the stationary cars, by which we train the modules for stationary vehicle boundary line detection.  
    }
    \label{dataset_od}
\end{figure}

\begin{figure}[th!]
    \centering
    \includegraphics[width=0.9\linewidth]{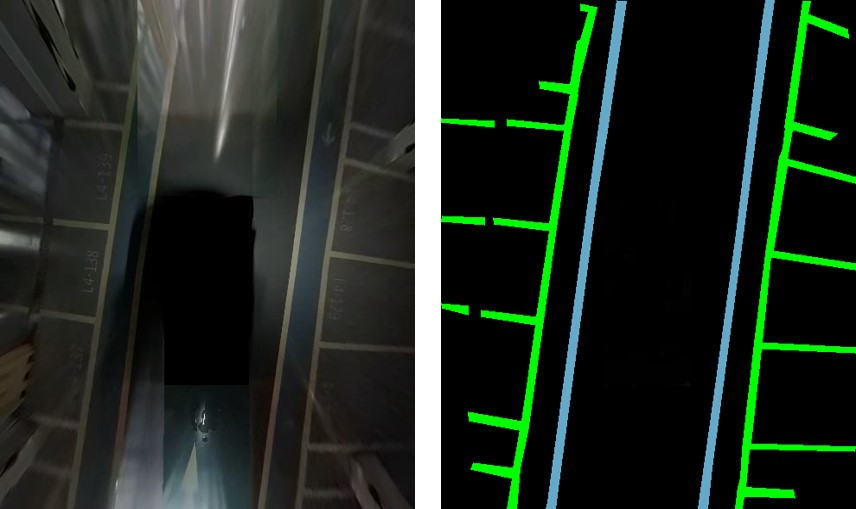}
    \caption{\textbf{An example of line landmark's labelling in BEV image.}
    We annotate the line landmark in the style of pixel-wise style.  
    }
    \label{dataset_seg}
\end{figure}

Different from the general detection benchmark in which each sample is a single image, 
we capture images using the surround-view camera system as detailed in Fig. \ref{ipm} where each sample consists of four fisheye images. 
And the image resolution is $1280\times 960$. 
To train our system, we annotate our dataset with two types of labels -- i.e., the bounding box of stationary vehicle Fig. \ref{dataset_od} and pixel-wise label map of line landmarks in BEV space as shown in Fig. \ref{dataset_seg}. 
\yz{Note that the stationary vehicle boundary line's detection depends on the vehicle boxes' key points (midpoints of the lower bottom), so the annotation for the stationary vehicle boundary line is equal to the annotation for the bounding box of stationary vehicles.
}

\subsubsection{Implementation details} 
We implement our system on the Qualcomm 820A platform with an Adreno 530 GPU and a Hexagon 680 DSP. 
The platform provides the computation power of 1.2 trillion operations per second (TOPS, \gan{Tera Operations Per Second}) which is weaker than the popular NVIDIA Xavier (40 TOPS) and Tesla's V3 Full self-driving computer claims (144 TOPS). Even though, our system achieves real-time line landmark detection.
\gan{In addition, we reveal the settings of hyper-parameters. We input the surround-view images with 960 pixels of $H_S$ and 1280 pixels of $W_S$.
For the BEV images, the $H_B$ is 600 and $W_B$ is 480.
The $h_m$ and $w_m$ contains four levels with 1/4, 1/8, 1/16 and 1/32 of the BEV image shape ($H_B$, $W_B$).  
We set the $\lambda_1$, $\lambda_2$ and $\lambda_3$ with 1.0, 1.0 and 1.0. 
}

\subsection{Performance of our line landmark detection system}
Our method achieves effective and efficient line landmark detection. We detail the performance in terms of accuracy and efficiency. 
\subsubsection{Accuracy}
We evaluate the accuracy of line landmark detection using false detection rate (FD) and missed detection rate (MD). We define FD as the ratio of incorrect detected line landmarks (e.g., line landmarks with the wrong line parameters.) and MD as the ratio of missing detection. FD and MD are critical metrics to measure the accuracy of line landmark detection systems. 
Typically, the valet parking system requires very low FD and MD.
We don't test on real distance error.
Real distance might be a good metric for some tasks. Nevertheless, in our case, we use the rate because we empirically observed that, in some cases, the real error distance could be very large, which dominates the final number and leads to a minor difference between methods.  
Thus, we choose to use the ratio as our metric.

As shown in Tab. \ref{tab:4lines_fd_md}, our method achieves the very low MD and FD in the test set. We achieve the low FD smaller than 3 percent and the lower MD smaller than 1 percent for all types of line landmarks. 
The very small MD means that we have a very high recall in terms of line landmark detection, which is critical to assure safety. 
For instance, we achieve nearly zero MD (0.09) to detect the stationary vehicle boundary, in which case, our system demonstrates very high accuracy in localizing the boundary between drivable and non-drivable areas. It's the key to guaranteeing collision-free valet parking. 

To further validate the effectiveness of our system, we adapt the existing works to our benchmark such that we have baselines for our method including UFLD \cite{qin2020ultra} and DMPR-PS \cite{huang2019dmpr}.  
Note that, unlike our method, those baselines don't aim for the detection of various line landmarks. We thus carefully adapt other methods to our benchmark.    
Nevertheless, all baselines, as shown in Tab. \ref{table_compare}, are still not applicable to line landmark detection. 
On the other hand, thanks to the carefully designed modules, our method achieves superior accuracy, \gan{where our multi-task architecture fully fuses the surround-view features and BEV features to produce better line detection with improved multi-task decoders and line fitting module.}  

\begin{table}[!ht]
    \centering
    \caption{The line landmark detection results with our dataset.}
    \label{tab:4lines_fd_md}
        \begin{threeparttable}
            \begin{tabular}{lll}
            \toprule 
                \textbf{Type}   & \textbf{FD (\%)} & \textbf{MD (\%)}\\
                \midrule
                Lane line                 &1.69   &0.17 \\
                Parking line               &2.83   &0.17 \\
                Median line                 &1.20   &0.13 \\
                Stationary vehicle boundary line  &1.07   &0.09 \\
            \bottomrule 
            \end{tabular}
        \end{threeparttable}
\end{table}
\begin{table}[!ht]
    \centering
    \footnotesize
    \setlength\tabcolsep{4pt} 
    \caption{Comparison in landmark detection using metrics of FD (\%) / MD (\%).}
    \label{table_compare}
        \begin{threeparttable}
            \begin{tabular}{lccc}
            \toprule 
                \textbf{Methods}  &\textbf{UFLD \cite{qin2020ultra}} & \textbf{DMPR-PS\cite{huang2019dmpr}} & \textbf{Ours}\\
                \midrule
                Lane line                &0.15/98.41  & -   & \textbf{1.69/0.17} \\
                Parking line             &0.17/99.51  &2.53/0.88   &\textbf{2.83/0.17} \\
                Median line               &1.02/97.74  &-   &\textbf{1.20/0.13} \\
                Stationary vehicle boundary line&-  & -   &\textbf{1.07/0.09} \\
            \bottomrule 
            \end{tabular}
        \end{threeparttable}
\end{table}

Specifically, we utilize UFLD \cite{qin2020ultra} to detect three physical lines by replacing the instance segmentation with semantic segmentation (i.e., same as line detection module). 
Regarding the DMPR-PS \cite{huang2019dmpr} -- a method for parking slot detection, we detect parking lines by applying our line fitting module after the parking slot detection. Experimental results show that, not surprisingly, our method achieves significantly better performance.        
It's noteworthy that, there are no existing works for stationary vehicle boundary line detection since they propose mental line is novel.    
Therefore, our method, with baselines being very worse or even inapplicable in line landmark detection, is significant to our research community given the importance of line landmark detection in valet parking. 

\begin{figure}[t!]
    \centering
    \includegraphics[width=1.0\linewidth]{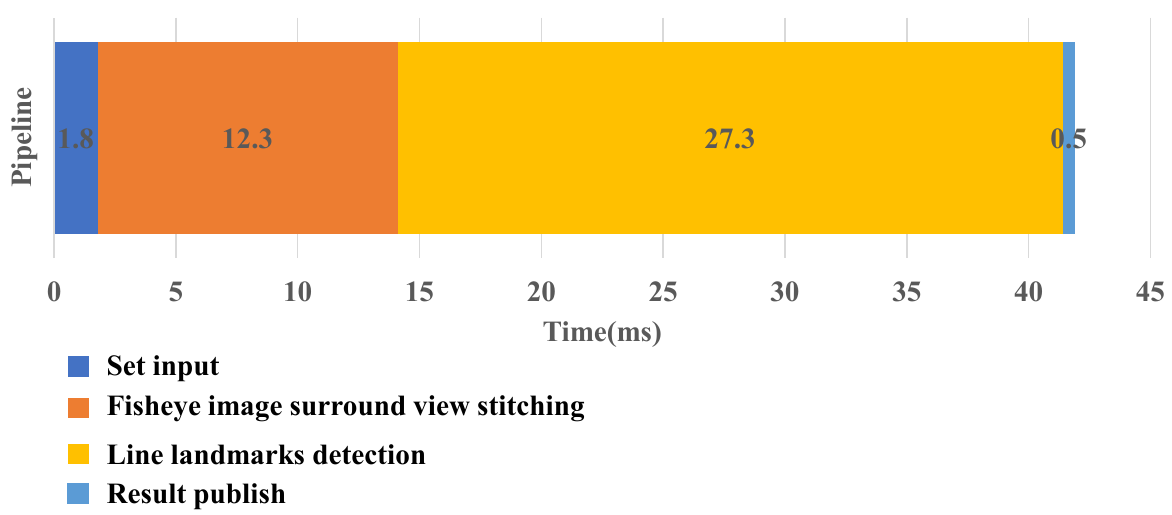}
    \caption{\textbf{Elapsed time of our system} on Qualcomm  820A  platform. We estimate the time for different time-consuming parts in our system. The total elapsed time of our system is less than 45 million seconds (ms) for each sample. Note that 60 ms is the minimum requirement for real-time detection.}
    \label{fig_time}
\end{figure}

\subsubsection{Efficiency}
Our line detection system is very efficient. We perform real-time detection, which is critical for valet parking. 
To quantify the efficiency, we estimate the elapsed time in the Qualcomm  820A  platform with weak computation ability. 

As shown in Fig. \ref{fig_time}, we estimate the elapsed time for the time-consuming parts in our system including the input pipeline (input images captured by four-way cameras, image stitching (i.e., BEV image generation),  line landmark detection, and result publish. 
The total elapsed time is less than 45 ms for each sample, which enables real-time detection where 60 ms/sample is the upper bound (minimum requirement) for time efficiency. It's noteworthy that our line landmark detection (i.e., our line landmark detection system) is only (27.3 ms/sample) since the process of images stitching is time-consuming.

\subsection{Ablation study}
We perform the ablation study to validate the core components of the system including multi-task architecture, transformer-based network, and our filtering backend.

\subsubsection{Multi-task Architecture}
\gan{
As shown in Tab. \ref{table_multitask_ablation}, the multi-task architecture achieves a better performance than the single-task architecture, where the multiple tasks are complementary to each other when we perform the tasks together. 
We achieve the robust fused features not only for the BEV-specific task (BEV semantic segmentation) but also for the image-feature-specific task (object detection).
}
Note that, for the sake of simplicity, we quantify the performance using accuracy (i.e., Accuracy = 1 - MD - FD).   

\begin{table}[!ht]
    \centering
    \caption{Accuracy(\%) of single semantic segmentation task, single object detection task and multi tasks.}
    \label{table_multitask_ablation}
        \begin{threeparttable}
            \begin{tabular}{lccc}
            \toprule 
                Type of line landmarks &SS & OD & Multi tasks\\
            \midrule
                Lane line                &97.86  &-   &\textbf{98.14} \\
                Parking line             &96.59  &-   &\textbf{97.00} \\
                Median line               &97.92  &-   &\textbf{98.67} \\
                Stationary vehicle boundary line &-  &98.41   &\textbf{98.84} \\
            \bottomrule 
            \end{tabular}
            \begin{tablenotes}
            \footnotesize
                \item[*] Note SS is semantic segmentation, OD is object detection.
            \end{tablenotes}
        \end{threeparttable}
\end{table}

\subsubsection{Semantic Segmentation}
As shown in Tab. \ref{table_seg_ablation}, we compare different semantic segmentation decoder designs for line landmarks, and it reveals that our transformer-based structure for semantic segmentation task outperforms other variants including DeepLabV3+ \cite{chen2018encoder} and the recent Swin \cite{liu2021swin}.
Same as Tab. \ref{table_multitask_ablation}, we utilize the metric of accuracy to evaluate the performance. 
The result shows that our transformer-based network leads to the best line landmark detection, and validates the effectiveness of our hierarchical level graph reasoning transformer to enhance the global and local relation modeling, which contributes to the long and ambiguous line contents detection.

\begin{table}[!ht]
    \centering
    \caption{\WS{Accuracy(\%) of different semantic segmentation decoder designs. `Time' denotes the time consumption on the NVIDIA GeForce RTX 4080.}}
    \label{table_seg_ablation}
        \begin{threeparttable}
            \begin{tabular}{lcccc}
            \toprule 
                Method   &Lane line & Parking line & Median Line &Time(ms)\\
                \midrule
                DeepLabv3+ \cite{chen2018encoder}    &94.84  &94.19   &93.96  &$\sim$12\\
                Swin \cite{liu2021swin}                 &96.38  &96.21   &96.45 &$\sim$40\\
                Ours               &\textbf{97.86}  &\textbf{96.59}   &\textbf{97.92} & $\sim$48\\
            \bottomrule 
            \end{tabular}
        \end{threeparttable}
\end{table}

\WS{
To explore the effectiveness of our graph transformer, we conduct the experiment on the public Cityscapes dataset~\cite{cityscapes}, as shown in Tab.~\ref{cityscapes}. 
To suit the large-scale dataset, we change the larger backbone together with compared methods, and the performance validates our graph transformer's advancement.
}

\begin{table}[!ht]
    \centering
    \caption{\WS{The performance of methods on the public Cityscapes dataset.}}
    \label{cityscapes}
        \begin{threeparttable}
            \begin{tabular}{lccc}
            \toprule 
                Method  &Backbone &val mIoU &test mIoU\\
                \midrule
                DeepLabv3+ \cite{chen2018encoder} &ResNet-101  &79.3 &80.1 \\
                Swin \cite{liu2021swin}       &Swin-L      &82.3 &81.3\\
                Ours       &Swin-L   &\textbf{82.9} &\textbf{81.9}  \\
            \bottomrule 
            \end{tabular}
        \end{threeparttable}
\end{table}

\WS{
\subsubsection{Object Detection}
We also evaluate the object detection task (though it's an intermediate result) on the surround-view valet parking dataset FPD~\cite{wu2022surround}, where we select the IoU criterion of 0.7 for object detection metrics: Average Precision (AP) and Average Recall (AR), indicating $AP_{2D}$ and $AR_{2D}$.
As shown in Tab.~\ref{tab:baselines_result}, for anchor-based RetinaNet \cite{article3}, it receives better $AP_{2D}$, but our anchor-free CenterNet follows the center-based structure \cite{duan2019centernet} which directly predict 2D bbox center and dimension, which achieves well-matched $AP_{2D}$ and better $AR_{2D}$. 
}

\begin{table}[!ht]
\small
\caption{\WS{The performance of methods on the FPD dataset. Highest result is marked with \textcolor{red}{red}. `Time' denotes the time consumption on the NVIDIA GeForce RTX 4080.}}
\label{tab:baselines_result}
\centering
\begin{threeparttable}
\begin{tabular}{lcccc}
\toprule 
Method & $AP_{2D}$     & $AR_{2D}$   &Time(ms)      \\ \midrule
RetinaNet \cite{article3}           & \textcolor{red}{45.6}  & 43.6  &$\sim$7 \\
CenterNet(Ours) \cite{duan2019centernet}  & 45.3  & \textcolor{red}{45.4}  &$\sim$5\\
\bottomrule 
\end{tabular}
\end{threeparttable}
\vspace{-0.1in}
\end{table}

\subsubsection{Filtering Backend}

We ablate our \ws{filtering} backend. 
In order to benchmark the impact of the filtering backend, we calculate the slope/intercept error ($\delta C_0$ and $\delta C_1$) of a sequence of samples where the error is defined as the offset between prediction and ground-truth parameters. 
\begin{figure}[!ht]
    \centering
    \includegraphics[width=\linewidth]{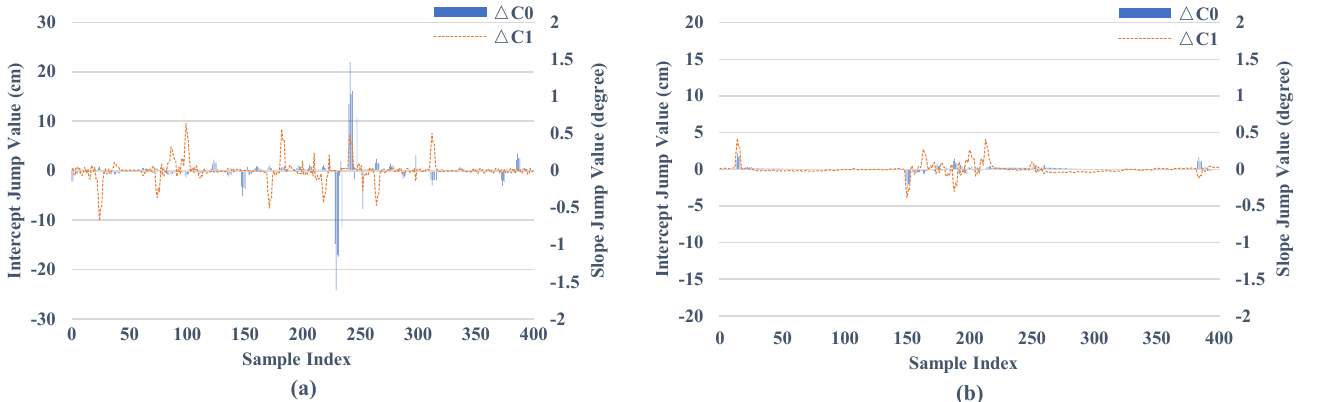}
    \caption{
    \textbf{Effectiveness of filtering backend}:  we show, with our filtering backend (b), our system achieves \WS{a lower error} than the system without filtering backend (a).
    }
    \label{Stability_output_experiment}
\end{figure}

As illustrated in Fig.~\ref{Stability_output_experiment}, we sequentially sample 400 frames and plot the error curves. 
We show that, through considering both temporal and multi-view consistency, our system effectively filters out the outlier detection with large errors. 
More importantly, we observe that, with our novel filtering backend, our system achieves a more stable and reliable line landmark detection given a sequence of surround-view frames.
Note that, while achieving better performance, our filtering backend is still very efficient as evidenced in our efficiency experiments.    

\begin{figure}[!b]
    \centering
    \includegraphics[width=0.7\linewidth]{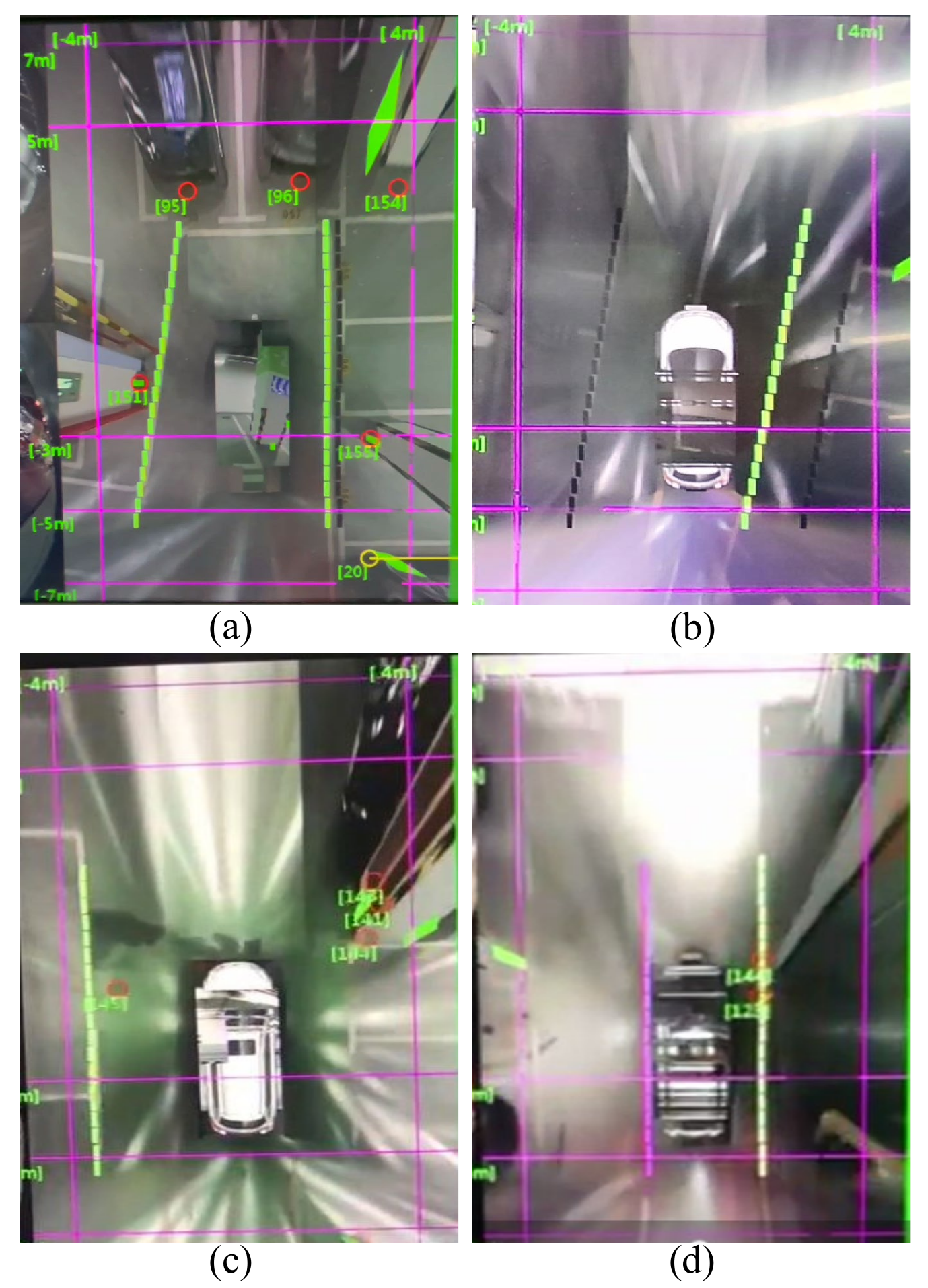}
    \caption{
    \yz{
    \textbf{Failure cases -- }
    (a) Line landmarks are perpendicular to the vehicle's heading direction; (b) median line under \WS{poor light conditions}; (c) parking line and stationary vehicle's boundary line under \WS{poor light conditions}; (d) lane line under \WS{poor light conditions}.
    }
    }
    \label{failure_case}
\end{figure}

\subsection{Failure cases}
Fig. \ref{failure_case} demonstrates some failure cases. 
Specifically, Fig. \ref{failure_case} (a) illustrates that our method fails when line landmarks are perpendicular to the vehicle's heading direction.
As shown in Fig. \ref{failure_case} (b), (c), and (d), 
our method fails to detect the median line, parking line, lane line, and stationary vehicle's boundary line under extremely poor light conditions (e.g., dark environments).
\gan{Our model doesn't work well under some special scenes like the above perpendicular condition and the poor light conditions, thus in the future we will collect more extreme pool conditions to improve the dataset's diversity and design more clever blocks to avoid failure cases.
In addition, in future work, we will upgrade the fitting algorithm to save more computation and consider more special scenes.}

\section{DISCUSSION}
In this section, we will systematize the experimental results and analyze the causes.
(i) For accuracy experiments, we exhibit our method's performance (Tab.~\ref{tab:4lines_fd_md}) and its comparison with existing works (Tab.~\ref{table_compare}) on our benchmark.
Our approach achieves pretty great accuracy, especially the low missed detection rate (MD), while the existing works reveal inapplicable in our line landmark detection, thus receiving sub-optimal performance.
(ii) For efficiency, we detail the elapsed time of our system on the Qualcomm 820A platform (Fig.10), which is less than 45 million seconds for each sample. 
Note that 60 ms is the minimum requirement for real-time detection.
(iii) For the multi-task architecture ablation study (Tab.~\ref{table_multitask_ablation}), we find that the multi-task architecture achieves more robust performance than the single-task one since the multiple tasks can promote each other by sharing helpful and meaningful features.
(iv) For the semantic segmentation task, we compare our approach with other existing approaches in Tab.~\ref{table_seg_ablation}. 
Our method harvests the best line landmark detection, for our hierarchical level graph reasoning transformer enhances the global and local relation modeling.
In addition, in public dataset evaluation (Tab.~\ref{cityscapes}), our approach still can make an advantage. 
(v) For the object detection task, we evaluate our center-based structure with the anchor-based RetinaNet (Tab.~\ref{tab:baselines_result}), which remains a well-matched performance with less time consumption.
(vi) The comparison with and without filtering (Fig.~\ref{Stability_output_experiment}) succeeds in confirming our filtering backend's effectiveness, where the offset is obviously inhibited with our beneficial filtering for multi-view consistency and temporal consistency.

In total, our method fulfills the superiority on both accuracy and efficiency compared with other existing approaches, which contributes to our elaborate designs, i.e. the multi-task architecture, the graph transformer for semantic segmentation task, the center-based CenterNet for object detection task and filtering backend for multi-view and temporal consistency, where our ablation studies have certified their effectiveness.

\section{CONCLUSIONS}

In this work, we present an effective yet efficient line landmark detection system for valet parking. 
To validate our method, we further propose a benchmark where we define four types of line landmarks, serving as the cornerstone for self-driving in a variety of parking lots.
Experimental results show that our system achieves low error rates of detection while being real-time on a low-cost computing platform.   

While we achieve the promising result, the error rate is still not zero whereas the perfect performance is critical for realistic valet parking.  
Thus, in the future, we would like to further improve modules in our system -- e.g., we further improve the generalization ability via enforcing inductive bias. 
Additionally, we would like to extend our method for generic detection -- apart from line landmarks, other landmarks would be also important such as parking slots, pedestrians, and road curbs to name a few. 
Note that it's non-trivial for those extensions even detection has been very successful in autonomous driving since, as discussed aforementioned, visual perception is much more challenging in valet parking environments than in autonomous driving areas.









\bibliography{sn-bibliography}


\begin{thebibliography}{55}
\ifx \bisbn   \undefined \def \bisbn  #1{ISBN #1}\fi
\ifx \binits  \undefined \def \binits#1{#1}\fi
\ifx \bauthor  \undefined \def \bauthor#1{#1}\fi
\ifx \batitle  \undefined \def \batitle#1{#1}\fi
\ifx \bjtitle  \undefined \def \bjtitle#1{#1}\fi
\ifx \bvolume  \undefined \def \bvolume#1{\textbf{#1}}\fi
\ifx \byear  \undefined \def \byear#1{#1}\fi
\ifx \bissue  \undefined \def \bissue#1{#1}\fi
\ifx \bfpage  \undefined \def \bfpage#1{#1}\fi
\ifx \blpage  \undefined \def \blpage #1{#1}\fi
\ifx \burl  \undefined \def \burl#1{\textsf{#1}}\fi
\ifx \doiurl  \undefined \def \doiurl#1{\url{https://doi.org/#1}}\fi
\ifx \betal  \undefined \def \betal{\textit{et al.}}\fi
\ifx \binstitute  \undefined \def \binstitute#1{#1}\fi
\ifx \binstitutionaled  \undefined \def \binstitutionaled#1{#1}\fi
\ifx \bctitle  \undefined \def \bctitle#1{#1}\fi
\ifx \beditor  \undefined \def \beditor#1{#1}\fi
\ifx \bpublisher  \undefined \def \bpublisher#1{#1}\fi
\ifx \bbtitle  \undefined \def \bbtitle#1{#1}\fi
\ifx \bedition  \undefined \def \bedition#1{#1}\fi
\ifx \bseriesno  \undefined \def \bseriesno#1{#1}\fi
\ifx \blocation  \undefined \def \blocation#1{#1}\fi
\ifx \bsertitle  \undefined \def \bsertitle#1{#1}\fi
\ifx \bsnm \undefined \def \bsnm#1{#1}\fi
\ifx \bsuffix \undefined \def \bsuffix#1{#1}\fi
\ifx \bparticle \undefined \def \bparticle#1{#1}\fi
\ifx \barticle \undefined \def \barticle#1{#1}\fi
\bibcommenthead
\ifx \bconfdate \undefined \def \bconfdate #1{#1}\fi
\ifx \botherref \undefined \def \botherref #1{#1}\fi
\ifx \url \undefined \def \url#1{\textsf{#1}}\fi
\ifx \bchapter \undefined \def \bchapter#1{#1}\fi
\ifx \bbook \undefined \def \bbook#1{#1}\fi
\ifx \bcomment \undefined \def \bcomment#1{#1}\fi
\ifx \oauthor \undefined \def \oauthor#1{#1}\fi
\ifx \citeauthoryear \undefined \def \citeauthoryear#1{#1}\fi
\ifx \endbibitem  \undefined \def \endbibitem {}\fi
\ifx \bconflocation  \undefined \def \bconflocation#1{#1}\fi
\ifx \arxivurl  \undefined \def \arxivurl#1{\textsf{#1}}\fi
\csname PreBibitemsHook\endcsname

\bibitem{song2019apollocar}
\begin{bchapter}
\bauthor{\bsnm{Song}, \binits{X.}},
\bauthor{\bsnm{Wang}, \binits{P.}},
\bauthor{\bsnm{Zhou}, \binits{D.}},
\bauthor{\bsnm{Zhu}, \binits{R.}},
\bauthor{\bsnm{Guan}, \binits{C.}},
\bauthor{\bsnm{Dai}, \binits{Y.}},
\bauthor{\bsnm{Su}, \binits{H.}},
\bauthor{\bsnm{Li}, \binits{H.}},
\bauthor{\bsnm{Yang}, \binits{R.}}:
\bctitle{Apollocar3d: A large 3d car instance understanding benchmark for autonomous driving}.
In: \bbtitle{Conference on Computer Vision and Pattern Recognition},
pp. \bfpage{5452}--\blpage{5462}
(\byear{2019}).
\bcomment{IEEE}
\end{bchapter}
\endbibitem

\bibitem{zhou2020joint}
\begin{bchapter}
\bauthor{\bsnm{Zhou}, \binits{D.}},
\bauthor{\bsnm{Fang}, \binits{J.}},
\bauthor{\bsnm{Song}, \binits{X.}},
\bauthor{\bsnm{Liu}, \binits{L.}},
\bauthor{\bsnm{Yin}, \binits{J.}},
\bauthor{\bsnm{Dai}, \binits{Y.}},
\bauthor{\bsnm{Li}, \binits{H.}},
\bauthor{\bsnm{Yang}, \binits{R.}}:
\bctitle{Joint 3d instance segmentation and object detection for autonomous driving}.
In: \bbtitle{Conference on Computer Vision and Pattern Recognition},
pp. \bfpage{1839}--\blpage{1849}
(\byear{2020})
\end{bchapter}
\endbibitem

\bibitem{wu2020motionnet}
\begin{bchapter}
\bauthor{\bsnm{Wu}, \binits{P.}},
\bauthor{\bsnm{Chen}, \binits{S.}},
\bauthor{\bsnm{Metaxas}, \binits{D.N.}}:
\bctitle{Motionnet: Joint perception and motion prediction for autonomous driving based on bird's eye view maps}.
In: \bbtitle{Conference on Computer Vision and Pattern Recognition},
pp. \bfpage{11385}--\blpage{11395}
(\byear{2020}).
\bcomment{IEEE}
\end{bchapter}
\endbibitem

\bibitem{lee2017real}
\begin{bchapter}
\bauthor{\bsnm{Lee}, \binits{D.-K.}},
\bauthor{\bsnm{Shin}, \binits{J.-S.}},
\bauthor{\bsnm{Jung}, \binits{J.-H.}},
\bauthor{\bsnm{Park}, \binits{S.-J.}},
\bauthor{\bsnm{Oh}, \binits{S.-J.}},
\bauthor{\bsnm{Lee}, \binits{I.-S.}}:
\bctitle{Real-time lane detection and tracking system using simple filter and kalman filter}.
In: \bbtitle{International Conference on Ubiquitous and Future Networks},
pp. \bfpage{275}--\blpage{277}
(\byear{2017}).
\bcomment{IEEE}
\end{bchapter}
\endbibitem

\bibitem{Lee2022robust}
\begin{bchapter}
\bauthor{\bsnm{Lee}, \binits{M.}},
\bauthor{\bsnm{Lee}, \binits{J.}},
\bauthor{\bsnm{Lee}, \binits{D.}},
\bauthor{\bsnm{Kim}, \binits{W.}},
\bauthor{\bsnm{Hwang}, \binits{S.}},
\bauthor{\bsnm{Lee}, \binits{S.}}:
\bctitle{Robust lane detection via expanded self attention}.
In: \bbtitle{Proceedings of the IEEE/CVF Winter Conference on Applications of Computer Vision},
pp. \bfpage{533}--\blpage{542}
(\byear{2022})
\end{bchapter}
\endbibitem

\bibitem{hu2022sim}
\begin{bchapter}
\bauthor{\bsnm{Hu}, \binits{C.}},
\bauthor{\bsnm{Hudson}, \binits{S.}},
\bauthor{\bsnm{Ethier}, \binits{M.}},
\bauthor{\bsnm{Al-Sharman}, \binits{M.}},
\bauthor{\bsnm{Rayside}, \binits{D.}},
\bauthor{\bsnm{Melek}, \binits{W.}}:
\bctitle{Sim-to-real domain adaptation for lane detection and classification in autonomous driving}.
In: \bbtitle{2022 IEEE Intelligent Vehicles Symposium (IV)},
pp. \bfpage{457}--\blpage{463}
(\byear{2022}).
\bcomment{IEEE}
\end{bchapter}
\endbibitem

\bibitem{chen2020collaborative}
\begin{bchapter}
\bauthor{\bsnm{Chen}, \binits{S.}},
\bauthor{\bsnm{Zhang}, \binits{N.}},
\bauthor{\bsnm{Sun}, \binits{H.}}:
\bctitle{Collaborative localization based on traffic landmarks for autonomous driving}.
In: \bbtitle{International Symposium on Circuits and Systems},
pp. \bfpage{1}--\blpage{5}
(\byear{2020}).
\bcomment{IEEE}
\end{bchapter}
\endbibitem

\bibitem{wu2020psdet}
\begin{bchapter}
\bauthor{\bsnm{Wu}, \binits{Z.}},
\bauthor{\bsnm{Sun}, \binits{W.}},
\bauthor{\bsnm{Wang}, \binits{M.}},
\bauthor{\bsnm{Wang}, \binits{X.}},
\bauthor{\bsnm{Ding}, \binits{L.}},
\bauthor{\bsnm{Wang}, \binits{F.}}:
\bctitle{Psdet: Efficient and universal parking slot detection}.
In: \bbtitle{Intelligent Vehicles Symposium},
pp. \bfpage{290}--\blpage{297}
(\byear{2020}).
\bcomment{IEEE}
\end{bchapter}
\endbibitem

\bibitem{qin2022ultra}
\begin{botherref}
\oauthor{\bsnm{Qin}, \binits{Z.}},
\oauthor{\bsnm{Zhang}, \binits{P.}},
\oauthor{\bsnm{Li}, \binits{X.}}:
Ultra fast deep lane detection with hybrid anchor driven ordinal classification.
IEEE Transactions on Pattern Analysis and Machine Intelligence
(2022)
\end{botherref}
\endbibitem

\bibitem{wang2022keypoint}
\begin{bchapter}
\bauthor{\bsnm{Wang}, \binits{J.}},
\bauthor{\bsnm{Ma}, \binits{Y.}},
\bauthor{\bsnm{Huang}, \binits{S.}},
\bauthor{\bsnm{Hui}, \binits{T.}},
\bauthor{\bsnm{Wang}, \binits{F.}},
\bauthor{\bsnm{Qian}, \binits{C.}},
\bauthor{\bsnm{Zhang}, \binits{T.}}:
\bctitle{A keypoint-based global association network for lane detection}.
In: \bbtitle{Proceedings of the IEEE/CVF Conference on Computer Vision and Pattern Recognition},
pp. \bfpage{1392}--\blpage{1401}
(\byear{2022})
\end{bchapter}
\endbibitem

\bibitem{feng2022rethinking}
\begin{bchapter}
\bauthor{\bsnm{Feng}, \binits{Z.}},
\bauthor{\bsnm{Guo}, \binits{S.}},
\bauthor{\bsnm{Tan}, \binits{X.}},
\bauthor{\bsnm{Xu}, \binits{K.}},
\bauthor{\bsnm{Wang}, \binits{M.}},
\bauthor{\bsnm{Ma}, \binits{L.}}:
\bctitle{Rethinking efficient lane detection via curve modeling}.
In: \bbtitle{Proceedings of the IEEE/CVF Conference on Computer Vision and Pattern Recognition},
pp. \bfpage{17062}--\blpage{17070}
(\byear{2022})
\end{bchapter}
\endbibitem

\bibitem{zheng2022clrnet}
\begin{bchapter}
\bauthor{\bsnm{Zheng}, \binits{T.}},
\bauthor{\bsnm{Huang}, \binits{Y.}},
\bauthor{\bsnm{Liu}, \binits{Y.}},
\bauthor{\bsnm{Tang}, \binits{W.}},
\bauthor{\bsnm{Yang}, \binits{Z.}},
\bauthor{\bsnm{Cai}, \binits{D.}},
\bauthor{\bsnm{He}, \binits{X.}}:
\bctitle{Clrnet: Cross layer refinement network for lane detection}.
In: \bbtitle{Proceedings of the IEEE/CVF Conference on Computer Vision and Pattern Recognition},
pp. \bfpage{898}--\blpage{907}
(\byear{2022})
\end{bchapter}
\endbibitem

\bibitem{pauls2021automatic}
\begin{bchapter}
\bauthor{\bsnm{Pauls}, \binits{J.-H.}},
\bauthor{\bsnm{Schmidt}, \binits{B.}},
\bauthor{\bsnm{Stiller}, \binits{C.}}:
\bctitle{Automatic mapping of tailored landmark representations for automated driving and map learning}.
In: \bbtitle{International Conference on Robotics and Automation},
pp. \bfpage{6725}--\blpage{6731}
(\byear{2021}).
\bcomment{IEEE}
\end{bchapter}
\endbibitem

\bibitem{wang2017landmarks}
\begin{bchapter}
\bauthor{\bsnm{Wang}, \binits{B.}},
\bauthor{\bsnm{Stafford-Fraser}, \binits{Q.}},
\bauthor{\bsnm{Robinson}, \binits{P.}},
\bauthor{\bsnm{Dias}, \binits{E.}},
\bauthor{\bsnm{Skrypchuk}, \binits{L.}}:
\bctitle{Landmarks based human-like guidance for driving navigation in an urban environment}.
In: \bbtitle{International Conference on Intelligent Transportation Systems},
pp. \bfpage{1}--\blpage{6}
(\byear{2017}).
\bcomment{IEEE}
\end{bchapter}
\endbibitem

\bibitem{liu2021condlanenet}
\begin{bchapter}
\bauthor{\bsnm{Liu}, \binits{L.}},
\bauthor{\bsnm{Chen}, \binits{X.}},
\bauthor{\bsnm{Zhu}, \binits{S.}},
\bauthor{\bsnm{Tan}, \binits{P.}}:
\bctitle{Condlanenet: a top-to-down lane detection framework based on conditional convolution}.
In: \bbtitle{Proceedings of the IEEE/CVF International Conference on Computer Vision},
pp. \bfpage{3773}--\blpage{3782}
(\byear{2021})
\end{bchapter}
\endbibitem

\bibitem{tabelini2021keep}
\begin{bchapter}
\bauthor{\bsnm{Tabelini}, \binits{L.}},
\bauthor{\bsnm{Berriel}, \binits{R.}},
\bauthor{\bsnm{Paixao}, \binits{T.M.}},
\bauthor{\bsnm{Badue}, \binits{C.}},
\bauthor{\bsnm{De~Souza}, \binits{A.F.}},
\bauthor{\bsnm{Oliveira-Santos}, \binits{T.}}:
\bctitle{Keep your eyes on the lane: Real-time attention-guided lane detection}.
In: \bbtitle{Proceedings of the IEEE/CVF Conference on Computer Vision and Pattern Recognition},
pp. \bfpage{294}--\blpage{302}
(\byear{2021})
\end{bchapter}
\endbibitem

\bibitem{2018IVreal}
\begin{bchapter}
\bauthor{\bsnm{Baek}, \binits{I.}},
\bauthor{\bsnm{Davies}, \binits{A.}},
\bauthor{\bsnm{Yan}, \binits{G.}},
\bauthor{\bsnm{Rajkumar}, \binits{R.R.}}:
\bctitle{Real-time detection, tracking, and classification of moving and stationary objects using multiple fisheye images}.
In: \bbtitle{Intelligent Vehicles Symposium},
pp. \bfpage{447}--\blpage{452}
(\byear{2018}).
\bcomment{IEEE}
\end{bchapter}
\endbibitem

\bibitem{wu2021disentangling}
\begin{bchapter}
\bauthor{\bsnm{Wu}, \binits{Z.}},
\bauthor{\bsnm{Zhang}, \binits{W.}},
\bauthor{\bsnm{Wang}, \binits{J.}},
\bauthor{\bsnm{Wang}, \binits{M.}},
\bauthor{\bsnm{Gan}, \binits{Y.}},
\bauthor{\bsnm{Gou}, \binits{X.}},
\bauthor{\bsnm{Fang}, \binits{M.}},
\bauthor{\bsnm{Song}, \binits{J.}}:
\bctitle{Disentangling and vectorization: A 3d visual perception approach for autonomous driving based on surround-view fisheye cameras}.
In: \bbtitle{IEEE/RJS International Conference on Intelligent RObots and Systems},
pp. \bfpage{5576}--\blpage{5582}
(\byear{2021}).
\bcomment{IEEE}
\end{bchapter}
\endbibitem

\bibitem{2018IROSend}
\begin{bchapter}
\bauthor{\bsnm{Toromanoff}, \binits{M.}},
\bauthor{\bsnm{Wirbel}, \binits{E.}},
\bauthor{\bsnm{Wilhelm}, \binits{F.}},
\bauthor{\bsnm{Vejarano}, \binits{C.}},
\bauthor{\bsnm{Perrotton}, \binits{X.}},
\bauthor{\bsnm{Moutarde}, \binits{F.}}:
\bctitle{End to end vehicle lateral control using a single fisheye camera}.
In: \bbtitle{IEEE/RJS International Conference on Intelligent RObots and Systems},
pp. \bfpage{3613}--\blpage{3619}
(\byear{2018}).
\bcomment{IEEE}
\end{bchapter}
\endbibitem

\bibitem{2021arxivomnidet}
\begin{barticle}
\bauthor{\bsnm{Kumar}, \binits{V.R.}},
\bauthor{\bsnm{Yogamani}, \binits{S.}},
\bauthor{\bsnm{Rashed}, \binits{H.}},
\bauthor{\bsnm{Sitsu}, \binits{G.}},
\bauthor{\bsnm{Witt}, \binits{C.}},
\bauthor{\bsnm{Leang}, \binits{I.}},
\bauthor{\bsnm{Milz}, \binits{S.}},
\bauthor{\bsnm{M{\"a}der}, \binits{P.}}:
\batitle{Omnidet: Surround view cameras based multi-task visual perception network for autonomous driving}.
\bjtitle{IEEE Robotics and Automation Letters}
\bvolume{6}(\bissue{2}),
\bfpage{2830}--\blpage{2837}
(\byear{2021})
\end{barticle}
\endbibitem

\bibitem{zhang2021avp}
\begin{bchapter}
\bauthor{\bsnm{Zhang}, \binits{C.}},
\bauthor{\bsnm{Liu}, \binits{H.}},
\bauthor{\bsnm{Xie}, \binits{Z.}},
\bauthor{\bsnm{Yang}, \binits{K.}},
\bauthor{\bsnm{Guo}, \binits{K.}},
\bauthor{\bsnm{Cai}, \binits{R.}},
\bauthor{\bsnm{Li}, \binits{Z.}}:
\bctitle{Avp-loc: Surround view localization and relocalization based on hd vector map for automated valet parking}.
In: \bbtitle{IEEE/RJS International Conference on Intelligent RObots and Systems},
pp. \bfpage{5552}--\blpage{5559}
(\byear{2021}).
\bcomment{IEEE}
\end{bchapter}
\endbibitem

\bibitem{kumar2020fisheyedistancenet}
\begin{bchapter}
\bauthor{\bsnm{Kumar}, \binits{V.R.}},
\bauthor{\bsnm{Hiremath}, \binits{S.A.}}, \betal:
\bctitle{Fisheyedistancenet: Self-supervised scale-aware distance estimation using monocular fisheye camera for autonomous driving}.
In: \bbtitle{International Conference on Robotics and Automation},
pp. \bfpage{574}--\blpage{581}
(\byear{2020}).
\bcomment{IEEE}
\end{bchapter}
\endbibitem

\bibitem{yahiaoui2019fisheyemodnet}
\begin{bchapter}
\bauthor{\bsnm{Yahiaoui}, \binits{M.}},
\bauthor{\bsnm{Rashed}, \binits{H.}},
\bauthor{\bsnm{Mariotti}, \binits{L.}},
\bauthor{\bsnm{Sistu}, \binits{G.}},
\bauthor{\bsnm{Clancy}, \binits{I.}},
\bauthor{\bsnm{Yahiaoui}, \binits{L.}},
\bauthor{\bsnm{Kumar}, \binits{V.R.}},
\bauthor{\bsnm{Yogamani}, \binits{S.}}:
\bctitle{Fisheyemodnet: Moving object detection on surround-view cameras for autonomous driving}.
In: \bbtitle{International Conference on Computer Vision Workshop}
(\byear{2019})
\end{bchapter}
\endbibitem

\bibitem{2020IROSAccuratelow}
\begin{bchapter}
\bauthor{\bsnm{Strobel}, \binits{K.}},
\bauthor{\bsnm{Zhu}, \binits{S.}},
\bauthor{\bsnm{Chang}, \binits{R.}},
\bauthor{\bsnm{Koppula}, \binits{S.}}:
\bctitle{Accurate, low-latency visual perception for autonomous racing: Challenges, mechanisms, and practical solutions}.
In: \bbtitle{2020 IEEE/RSJ International Conference on Intelligent Robots and Systems (IROS)},
pp. \bfpage{1969}--\blpage{1975}
(\byear{2020}).
\bcomment{IEEE}
\end{bchapter}
\endbibitem

\bibitem{qin2020avp}
\begin{bchapter}
\bauthor{\bsnm{Qin}, \binits{T.}},
\bauthor{\bsnm{Chen}, \binits{T.}},
\bauthor{\bsnm{Chen}, \binits{Y.}},
\bauthor{\bsnm{Su}, \binits{Q.}}:
\bctitle{Avp-slam: Semantic visual mapping and localization for autonomous vehicles in the parking lot}.
In: \bbtitle{IEEE/RJS International Conference on Intelligent RObots and Systems},
pp. \bfpage{5939}--\blpage{5945}
(\byear{2020}).
\bcomment{IEEE}
\end{bchapter}
\endbibitem

\bibitem{ouyang2020online}
\begin{bchapter}
\bauthor{\bsnm{Ouyang}, \binits{Z.}},
\bauthor{\bsnm{Hu}, \binits{L.}},
\bauthor{\bsnm{Lu}, \binits{Y.}},
\bauthor{\bsnm{Wang}, \binits{Z.}},
\bauthor{\bsnm{Peng}, \binits{X.}},
\bauthor{\bsnm{Kneip}, \binits{L.}}:
\bctitle{Online calibration of exterior orientations of a vehicle-mounted surround-view camera system}.
In: \bbtitle{International Conference on Robotics and Automation},
pp. \bfpage{4990}--\blpage{4996}
(\byear{2020}).
\bcomment{IEEE}
\end{bchapter}
\endbibitem

\bibitem{wang2020reliable}
\begin{bchapter}
\bauthor{\bsnm{Wang}, \binits{Y.}},
\bauthor{\bsnm{Huang}, \binits{K.}},
\bauthor{\bsnm{Peng}, \binits{X.}},
\bauthor{\bsnm{Li}, \binits{H.}},
\bauthor{\bsnm{Kneip}, \binits{L.}}:
\bctitle{Reliable frame-to-frame motion estimation for vehicle-mounted surround-view camera systems}.
In: \bbtitle{International Conference on Robotics and Automation},
pp. \bfpage{1660}--\blpage{1666}
(\byear{2020}).
\bcomment{IEEE}
\end{bchapter}
\endbibitem

\bibitem{2018ECCVWsemantic}
\begin{bchapter}
\bauthor{\bsnm{Blott}, \binits{G.}},
\bauthor{\bsnm{Takami}, \binits{M.}},
\bauthor{\bsnm{Heipke}, \binits{C.}}:
\bctitle{Semantic segmentation of fisheye images}.
In: \bbtitle{European Conference on Computer Vision Workshop},
pp. \bfpage{181}--\blpage{196}
(\byear{2018})
\end{bchapter}
\endbibitem

\bibitem{kumar2021syndistnet}
\begin{bchapter}
\bauthor{\bsnm{Kumar}, \binits{V.R.}},
\bauthor{\bsnm{Klingner}, \binits{M.}},
\bauthor{\bsnm{Yogamani}, \binits{S.}},
\bauthor{\bsnm{Milz}, \binits{S.}},
\bauthor{\bsnm{Fingscheidt}, \binits{T.}},
\bauthor{\bsnm{Mader}, \binits{P.}}:
\bctitle{Syndistnet: Self-supervised monocular fisheye camera distance estimation synergized with semantic segmentation for autonomous driving}.
In: \bbtitle{Winter Conference on Applications of Computer Vision},
pp. \bfpage{61}--\blpage{71}
(\byear{2021}).
\bcomment{IEEE}
\end{bchapter}
\endbibitem

\bibitem{ye2020universal}
\begin{bchapter}
\bauthor{\bsnm{Ye}, \binits{Y.}},
\bauthor{\bsnm{Yang}, \binits{K.}},
\bauthor{\bsnm{Xiang}, \binits{K.}},
\bauthor{\bsnm{Wang}, \binits{J.}},
\bauthor{\bsnm{Wang}, \binits{K.}}:
\bctitle{Universal semantic segmentation for fisheye urban driving images}.
In: \bbtitle{International Conference on Systems, Man, and Cybernetics},
pp. \bfpage{648}--\blpage{655}
(\byear{2020}).
\bcomment{IEEE}
\end{bchapter}
\endbibitem

\bibitem{zaffaroni2019estimation}
\begin{bchapter}
\bauthor{\bsnm{Zaffaroni}, \binits{M.}},
\bauthor{\bsnm{Grangetto}, \binits{M.}},
\bauthor{\bsnm{Farasin}, \binits{A.}}:
\bctitle{Estimation of speed and distance of surrounding vehicles from a single camera}.
In: \bbtitle{International Conference on Image Analysis and Processing},
pp. \bfpage{388}--\blpage{398}
(\byear{2019}).
\bcomment{Springer}
\end{bchapter}
\endbibitem

\bibitem{komatsu2020360}
\begin{bchapter}
\bauthor{\bsnm{Komatsu}, \binits{R.}},
\bauthor{\bsnm{Fujii}, \binits{H.}},
\bauthor{\bsnm{Tamura}, \binits{Y.}},
\bauthor{\bsnm{Yamashita}, \binits{A.}},
\bauthor{\bsnm{Asama}, \binits{H.}}:
\bctitle{360 depth estimation from multiple fisheye images with origami crown representation of icosahedron}.
In: \bbtitle{IEEE/RJS International Conference on Intelligent RObots and Systems},
pp. \bfpage{10092}--\blpage{10099}
(\byear{2020}).
\bcomment{IEEE}
\end{bchapter}
\endbibitem

\bibitem{plaut2020monocular}
\begin{botherref}
\oauthor{\bsnm{Plaut}, \binits{E.}},
\oauthor{\bsnm{Yaacov}, \binits{E.B.}},
\oauthor{\bsnm{Shlomo}, \binits{B.E.}}:
Monocular 3d object detection in cylindrical images from fisheye cameras.
arXiv preprint arXiv:2003.03759
(2020)
\end{botherref}
\endbibitem

\bibitem{li2020fisheyedet}
\begin{barticle}
\bauthor{\bsnm{Li}, \binits{T.}},
\bauthor{\bsnm{Tong}, \binits{G.}},
\bauthor{\bsnm{Tang}, \binits{H.}},
\bauthor{\bsnm{Li}, \binits{B.}},
\bauthor{\bsnm{Chen}, \binits{B.}}:
\batitle{Fisheyedet: A self-study and contour-based object detector in fisheye images}.
\bjtitle{IEEE Access}
\bvolume{8},
\bfpage{71739}--\blpage{71751}
(\byear{2020})
\end{barticle}
\endbibitem

\bibitem{2020xiao}
\begin{barticle}
\bauthor{\bsnm{Xiao}, \binits{J.}},
\bauthor{\bsnm{Xiong}, \binits{W.}},
\bauthor{\bsnm{Yao}, \binits{Y.}},
\bauthor{\bsnm{Li}, \binits{L.}},
\bauthor{\bsnm{Klette}, \binits{R.}}:
\batitle{Lane detection algorithm based on road structure and extended kalman filter}.
\bjtitle{International Journal of Digital Crime and Forensics}
\bvolume{12}(\bissue{2}),
\bfpage{1}--\blpage{20}
(\byear{2020})
\end{barticle}
\endbibitem

\bibitem{hu2019mapping}
\begin{bchapter}
\bauthor{\bsnm{Hu}, \binits{J.}},
\bauthor{\bsnm{Yang}, \binits{M.}},
\bauthor{\bsnm{Xu}, \binits{H.}},
\bauthor{\bsnm{He}, \binits{Y.}},
\bauthor{\bsnm{Wang}, \binits{C.}}:
\bctitle{Mapping and localization using semantic road marking with centimeter-level accuracy in indoor parking lots}.
In: \bbtitle{International Conference on Intelligent Transportation Systems},
pp. \bfpage{4068}--\blpage{4073}
(\byear{2019}).
\bcomment{IEEE}
\end{bchapter}
\endbibitem

\bibitem{qin2018vins}
\begin{barticle}
\bauthor{\bsnm{Qin}, \binits{T.}},
\bauthor{\bsnm{Li}, \binits{P.}},
\bauthor{\bsnm{Shen}, \binits{S.}}:
\batitle{Vins-mono: A robust and versatile monocular visual-inertial state estimator}.
\bjtitle{IEEE Transactions on Robotics}
\bvolume{34}(\bissue{4}),
\bfpage{1004}--\blpage{1020}
(\byear{2018})
\end{barticle}
\endbibitem

\bibitem{2020Marzonggui}
\begin{barticle}
\bauthor{\bsnm{Marzougui}, \binits{M.}},
\bauthor{\bsnm{Alasiry}, \binits{A.}},
\bauthor{\bsnm{Kortli}, \binits{Y.}},
\bauthor{\bsnm{Baili}, \binits{J.}}:
\batitle{A lane tracking method based on progressive probabilistic hough transform}.
\bjtitle{IEEE access}
\bvolume{8},
\bfpage{84893}--\blpage{84905}
(\byear{2020})
\end{barticle}
\endbibitem

\bibitem{qin2020ultra}
\begin{bchapter}
\bauthor{\bsnm{Qin}, \binits{Z.}},
\bauthor{\bsnm{Wang}, \binits{H.}},
\bauthor{\bsnm{Li}, \binits{X.}}:
\bctitle{Ultra fast structure-aware deep lane detection}.
In: \bbtitle{European Conference on Computer Vision},
pp. \bfpage{276}--\blpage{291}
(\byear{2020}).
\bcomment{Springer}
\end{bchapter}
\endbibitem

\bibitem{huang2019dmpr}
\begin{bchapter}
\bauthor{\bsnm{Huang}, \binits{J.}},
\bauthor{\bsnm{Zhang}, \binits{L.}},
\bauthor{\bsnm{Shen}, \binits{Y.}},
\bauthor{\bsnm{Zhang}, \binits{H.}},
\bauthor{\bsnm{Zhao}, \binits{S.}},
\bauthor{\bsnm{Yang}, \binits{Y.}}:
\bctitle{Dmpr-ps: a novel approach for parking-slot detection using directional marking-point regression}.
In: \bbtitle{IEEE International Conference on Multimedia and Expo},
pp. \bfpage{212}--\blpage{217}
(\byear{2019}).
\bcomment{IEEE}
\end{bchapter}
\endbibitem

\bibitem{2018Song}
\begin{barticle}
\bauthor{\bsnm{Song}, \binits{M.}},
\bauthor{\bsnm{Kim}, \binits{C.}},
\bauthor{\bsnm{Kim}, \binits{M.}},
\bauthor{\bsnm{Yi}, \binits{K.}}:
\batitle{Robust lane tracking algorithm for forward target detection of automated driving vehicles}.
\bjtitle{Proceedings of the Institution of Mechanical Engineers, Part D: Journal of automobile engineering}
\bvolume{233}(\bissue{7}),
\bfpage{1930}--\blpage{1949}
(\byear{2019})
\end{barticle}
\endbibitem

\bibitem{zhang2000flexible}
\begin{barticle}
\bauthor{\bsnm{Zhang}, \binits{Z.}}:
\batitle{A flexible new technique for camera calibration}.
\bjtitle{IEEE Transactions on Pattern Analysis and Machine Intelligence}
\bvolume{22}(\bissue{11}),
\bfpage{1330}--\blpage{1334}
(\byear{2000})
\end{barticle}
\endbibitem

\bibitem{dla}
\begin{bchapter}
\bauthor{\bsnm{Yu}, \binits{F.}},
\bauthor{\bsnm{Wang}, \binits{D.}},
\bauthor{\bsnm{Shelhamer}, \binits{E.}},
\bauthor{\bsnm{Darrell}, \binits{T.}}:
\bctitle{Deep layer aggregation}.
In: \bbtitle{Computer Vision and Pattern Recognition},
pp. \bfpage{2403}--\blpage{2412}
(\byear{2018}).
\bcomment{IEEE}
\end{bchapter}
\endbibitem

\bibitem{bruna2014spectral}
\begin{bchapter}
\bauthor{\bsnm{Bruna}, \binits{J.}},
\bauthor{\bsnm{Zaremba}, \binits{W.}},
\bauthor{\bsnm{Szlam}, \binits{A.}},
\bauthor{\bsnm{LeCun}, \binits{Y.}}:
\bctitle{Spectral networks and locally connected networks on graphs}.
In: \bbtitle{International Conference on Learning Representations}
(\byear{2014}).
\bcomment{OpenReview.net}
\end{bchapter}
\endbibitem

\bibitem{liu2021swin}
\begin{bchapter}
\bauthor{\bsnm{Liu}, \binits{Z.}},
\bauthor{\bsnm{Lin}, \binits{Y.}},
\bauthor{\bsnm{Cao}, \binits{Y.}},
\bauthor{\bsnm{Hu}, \binits{H.}},
\bauthor{\bsnm{Wei}, \binits{Y.}},
\bauthor{\bsnm{Zhang}, \binits{Z.}},
\bauthor{\bsnm{Lin}, \binits{S.}},
\bauthor{\bsnm{Guo}, \binits{B.}}:
\bctitle{Swin transformer: Hierarchical vision transformer using shifted windows}.
In: \bbtitle{IEEE International Conference on Computer Vision},
pp. \bfpage{10012}--\blpage{10022}
(\byear{2021}).
\bcomment{IEEE}
\end{bchapter}
\endbibitem

\bibitem{2019arxivobjects}
\begin{botherref}
\oauthor{\bsnm{Zhou}, \binits{X.}},
\oauthor{\bsnm{Wang}, \binits{D.}},
\oauthor{\bsnm{Kr{\"a}henb{\"u}hl}, \binits{P.}}:
Objects as points.
arXiv preprint arXiv:1904.07850
(2019)
\end{botherref}
\endbibitem

\bibitem{mallot1991inverse}
\begin{barticle}
\bauthor{\bsnm{Mallot}, \binits{H.A.}},
\bauthor{\bsnm{B{\"u}lthoff}, \binits{H.H.}},
\bauthor{\bsnm{Little}, \binits{J.}},
\bauthor{\bsnm{Bohrer}, \binits{S.}}:
\batitle{Inverse perspective mapping simplifies optical flow computation and obstacle detection}.
\bjtitle{Biological Cybernetics}
\bvolume{64}(\bissue{3}),
\bfpage{177}--\blpage{185}
(\byear{1991})
\end{barticle}
\endbibitem

\bibitem{kalman1}
\begin{botherref}
\oauthor{\bsnm{Cen}, \binits{R.}},
\oauthor{\bsnm{Jiang}, \binits{T.}},
\oauthor{\bsnm{Tan}, \binits{Y.}},
\oauthor{\bsnm{Su}, \binits{X.}},
\oauthor{\bsnm{Xue}, \binits{F.}}:
A low-cost visual inertial odometry for mobile vehicle based on double stage kalman filter.
Signal Processing,
108537
(2022)
\end{botherref}
\endbibitem

\bibitem{kalman2}
\begin{bchapter}
\bauthor{\bsnm{Bloesch}, \binits{M.}},
\bauthor{\bsnm{Omari}, \binits{S.}},
\bauthor{\bsnm{Hutter}, \binits{M.}},
\bauthor{\bsnm{Siegwart}, \binits{R.}}:
\bctitle{Robust visual inertial odometry using a direct ekf-based approach}.
In: \bbtitle{IEEE/RJS International Conference on Intelligent RObots and Systems},
pp. \bfpage{298}--\blpage{304}
(\byear{2015}).
\bcomment{IEEE}
\end{bchapter}
\endbibitem

\bibitem{kalman3}
\begin{barticle}
\bauthor{\bsnm{Faessler}, \binits{M.}},
\bauthor{\bsnm{Fontana}, \binits{F.}},
\bauthor{\bsnm{Forster}, \binits{C.}},
\bauthor{\bsnm{Mueggler}, \binits{E.}},
\bauthor{\bsnm{Pizzoli}, \binits{M.}},
\bauthor{\bsnm{Scaramuzza}, \binits{D.}}:
\batitle{Autonomous, vision-based flight and live dense 3d mapping with a quadrotor micro aerial vehicle}.
\bjtitle{Journal of Field Robotics}
\bvolume{33}(\bissue{4}),
\bfpage{431}--\blpage{450}
(\byear{2016})
\end{barticle}
\endbibitem

\bibitem{chen2018encoder}
\begin{bchapter}
\bauthor{\bsnm{Chen}, \binits{L.-C.}},
\bauthor{\bsnm{Zhu}, \binits{Y.}},
\bauthor{\bsnm{Papandreou}, \binits{G.}},
\bauthor{\bsnm{Schroff}, \binits{F.}},
\bauthor{\bsnm{Adam}, \binits{H.}}:
\bctitle{Encoder-decoder with atrous separable convolution for semantic image segmentation}.
In: \bbtitle{European Conference on Computer Vision},
pp. \bfpage{833}--\blpage{851}
(\byear{2018}).
\bcomment{Springer}
\end{bchapter}
\endbibitem

\bibitem{cityscapes}
\begin{bchapter}
\bauthor{\bsnm{Cordts}, \binits{M.}},
\bauthor{\bsnm{Omran}, \binits{M.}},
\bauthor{\bsnm{Ramos}, \binits{S.}},
\bauthor{\bsnm{Rehfeld}, \binits{T.}},
\bauthor{\bsnm{Enzweiler}, \binits{M.}},
\bauthor{\bsnm{Benenson}, \binits{R.}},
\bauthor{\bsnm{Franke}, \binits{U.}},
\bauthor{\bsnm{Roth}, \binits{S.}},
\bauthor{\bsnm{Schiele}, \binits{B.}}:
\bctitle{The cityscapes dataset for semantic urban scene understanding}.
In: \bbtitle{Computer Vision and Pattern Recognition},
pp. \bfpage{3213}--\blpage{3223}
(\byear{2016}).
\bcomment{IEEE}
\end{bchapter}
\endbibitem

\bibitem{wu2022surround}
\begin{botherref}
\oauthor{\bsnm{Wu}, \binits{Z.}},
\oauthor{\bsnm{Gan}, \binits{Y.}},
\oauthor{\bsnm{Li}, \binits{X.}},
\oauthor{\bsnm{Wu}, \binits{Y.}},
\oauthor{\bsnm{Wang}, \binits{X.}},
\oauthor{\bsnm{Xu}, \binits{T.}},
\oauthor{\bsnm{Wang}, \binits{F.}}:
Surround-view fisheye bev-perception for valet parking: Dataset, baseline and distortion-insensitive multi-task framework.
IEEE Transactions on Intelligent Vehicles
(2022)
\end{botherref}
\endbibitem

\bibitem{article3}
\begin{bchapter}
\bauthor{\bsnm{Lin}, \binits{T.-Y.}},
\bauthor{\bsnm{Goyal}, \binits{P.}},
\bauthor{\bsnm{Girshick}, \binits{R.}},
\bauthor{\bsnm{He}, \binits{K.}},
\bauthor{\bsnm{Doll{\'a}r}, \binits{P.}}:
\bctitle{Focal loss for dense object detection}.
In: \bbtitle{IEEE Int. Conf. on Computer Vision (ICCV)},
pp. \bfpage{2980}--\blpage{2988}
(\byear{2017}).
\bcomment{IEEE}
\end{bchapter}
\endbibitem

\bibitem{duan2019centernet}
\begin{bchapter}
\bauthor{\bsnm{Duan}, \binits{K.}},
\bauthor{\bsnm{Bai}, \binits{S.}},
\bauthor{\bsnm{Xie}, \binits{L.}},
\bauthor{\bsnm{Qi}, \binits{H.}},
\bauthor{\bsnm{Huang}, \binits{Q.}},
\bauthor{\bsnm{Tian}, \binits{Q.}}:
\bctitle{Center{N}et: Keypoint triplets for object detection}.
In: \bbtitle{IEEE Int. Conf. on Computer Vision (ICCV)},
pp. \bfpage{6569}--\blpage{6578}
(\byear{2019}).
\bcomment{IEEE}
\end{bchapter}
\endbibitem

\end{thebibliography}


\end{document}